\pdfoutput=1

\documentclass[11pt]{article}

\usepackage{EMNLP2022}

\usepackage{times}
\usepackage{latexsym}
\usepackage[T1]{fontenc}
\usepackage[utf8]{inputenc}
\usepackage{microtype}
\usepackage{inconsolata}
\usepackage{times}
\usepackage{latexsym}
\usepackage{amsfonts}
\usepackage{amsmath}
\usepackage{amssymb}
\usepackage{multirow}
\usepackage{float}
\usepackage{booktabs}
\usepackage{fdsymbol}
\usepackage{xurl}

\usepackage{siunitx}

\makeatletter
\def\Url@twoslashes{\mathchar`\/\@ifnextchar/{\kern-.2em}{}}
\g@addto@macro\UrlSpecials{\do\/{\Url@twoslashes}}
\makeatother

\usepackage{stfloats}
\usepackage{flushend}

\usepackage{microtype}
\usepackage{graphicx}
\graphicspath{{./figures/}}
\usepackage{caption}
\usepackage{subcaption}
\usepackage[inline,shortlabels]{enumitem}
\usepackage[ruled,vlined]{algorithm2e}

\def \figurename {Fig.}
\def \figurenamelong {Figure}  

\def \Tablename {Table}
\def \appendixname {Appendix}

\newcommand{\figref}[1]{\figurename~\ref{#1}}
\newcommand{\Figref}[1]{\figurenamelong~\ref{#1}}
\newcommand{\tabref}[1]{\Tablename~\ref{#1}}
\newcommand{\secref}[1]{§\ref{#1}} 
\newcommand{\appref}[1]{\appendixname~\ref{#1}}

\newcommand{\eg}{e.g., }
\newcommand{\cf}{cf.\ }
\newcommand{\ie}{i.e., }
\newcommand{\vs}{vs.\ }

\def \counterfactual {\textsc{cad}}
\def \Dataset {\mathbf{\mathcal{D}}}
\def \Did {\Dataset_\textsc{id}}
\def \Dcf {\Dataset_\counterfactual}
\newcommand{\Dset}[2]{D_\textsc{#2}^\textit{#1}} 
\newcommand{\Dood}[1]{D_\textsc{ood}^\textsc{#1}} 
\def \Damzn {\Dood{amzn}}
\def \Dyelp {\Dood{yelp}}
\def \Dsemeval {\Dood{se}}  

\def \xc {x_\counterfactual}
\def \yc {y_\counterfactual}
\def \X {\mathcal{X}}
\def \Y {\mathcal{Y}}
\def \Rd {\mathbb{R}^d}
\def \Rdd {\mathbb{R}^{d\times d}}
\def \myplus {\scalebox{0.5}{$+$}}
\def \mymin {\scalebox{0.5}{$-$}}
\def \tp {t^{\myplus}}
\def \tn {t^{\mymin}}

\def \on {\Vec{o}_{\mymin}}

\def \Wn {W^{\mymin}}

\def \bn {b^{\mymin}}

\def \rn {r^{\mymin}}
\DeclareMathOperator*{\avg}{avg}

\usepackage{color}

\definecolor{positive}{RGB}{43, 102, 60}
\definecolor{negative}{RGB}{178, 54, 52}

\title{Robustifying Sentiment Classification\\ by Maximally Exploiting Few Counterfactuals}

\author{{\centering 
    Maarten De Raedt$^{\diamondsuit\clubsuit}$~ 
    Fréderic Godin$^\diamondsuit$~
    Chris Develder$^\clubsuit$~
    Thomas Demeester$^\clubsuit$}  \\
    $^\diamondsuit$ Sinch Chatlayer ~~$^\clubsuit$ Ghent University \\
    \texttt{\{maarten.deraedt, chris.develder, thomas.demeester\}@ugent.be}  \\
    \texttt{frederic.godin@sinch.com} \\
}

\begin{document}
\maketitle
\begin{abstract}
For text classification tasks, finetuned language models perform remarkably well. Yet, they tend to rely on spurious patterns in training data, thus limiting their performance on out-of-distribution (OOD) test data.
Among recent models aiming to avoid this spurious pattern problem, adding extra counterfactual samples to the training data has proven to be very effective.
Yet, counterfactual data generation is costly since it relies on human annotation.
Thus, we propose a novel solution that only requires annotation of a small fraction (\eg 1\%) of the original training data, and uses automatic generation of extra counterfactuals in an encoding vector space.
We demonstrate the effectiveness of our approach in sentiment classification, using IMDb data for training and other sets for OOD tests (\ie Amazon, SemEval and Yelp).
We achieve noticeable accuracy improvements by adding only 1\% manual counterfactuals: +3\% compared to adding +100\% in-distribution training samples, +1.3\% compared to alternate counterfactual approaches.
\end{abstract}

\section{Introduction and Related Work}
\label{sec:intro}

For a wide range of text classification tasks, finetuning large pretrained language models \cite{devlin-etal-2019-bert, liu2019roberta, clark-etal-2020-pre, lewis-etal-2020-bart} on task-specific data has been proven very effective.
Yet, analysis has shown that their predictions tend to rely on spurious patterns \cite{poliak-etal-2018-hypothesis, gururangan-etal-2018-annotation, kiritchenko-mohammad-2018-examining, mccoy-etal-2019-right, niven-kao-2019-probing, zmigrod-etal-2019-counterfactual, wang-culotta-2020-identifying}, \ie features that from a human perspective are not indicative for the classifier's label.
For instance, \citet{kaushik2019learning} found the rather neutral words ``will'', ``my'' and ``has'' to be important for a positive sentiment classification.
Such reliance on spurious patterns were suspected to degrade performance on out-of-distribution (OOD) test data, distributionally different from training data \cite{quinonero2008dataset}. 
Specifically for sentiment classification, this suspicion has been confirmed by \citet{kaushik2019learning, kaushik2020explaining}.

\begin{figure}[t]
    \centering
    \includegraphics[width=\columnwidth]{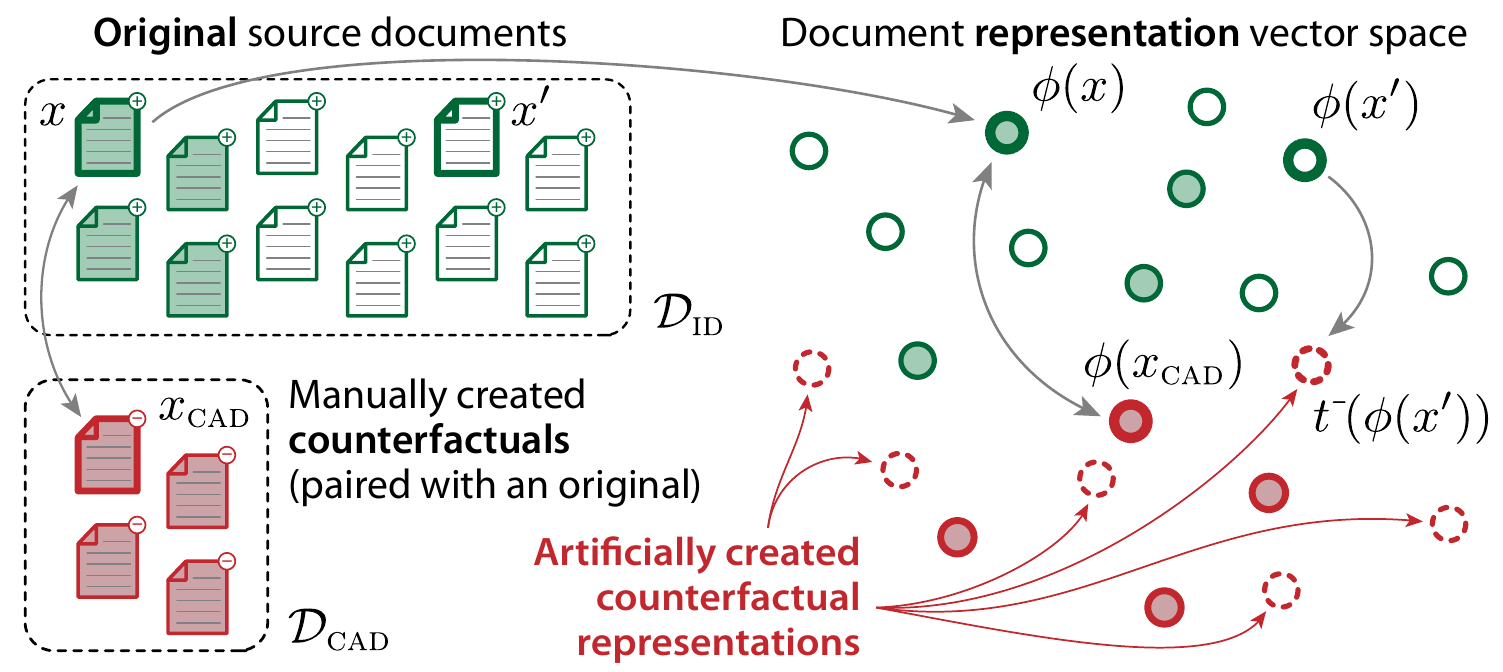}
    \caption{We propose to generate counterfactuals in representation space, learning --- from only a few manually created counterfactuals --- a mapping function $t$ to transform a document representation $\phi(x)$ to a counterfactual one (having the opposite classification label). Illustration for positively labeled originals only.}
    \label{fig:counterfactual-generation-idea}
\end{figure}

For mitigating the spurious pattern effect, generic methods include regularization of masked language models, which limits over-reliance on a limited set of keywords \cite{moon2021masker}. Alternatively, to improve robustness in imbalanced data settings, additional training samples can be automatically created \cite{han2021counterfactual}.
Other approaches rely on adding extra training data by \emph{human} annotation.
Specifically to avoid spurious patterns, \citet{kaushik2019learning} proposed Counterfactually Augmented Data (CAD), where annotators minimally revise training data to flip their labels: training on both original and counterfactual samples reduced spurious patterns.
Rather than editing existing samples, \citet{katakkar2021practical} propose to annotate them with text spans supporting the assigned labels as a ``rationale'' \cite{pruthi-etal-2020-weakly, jain-etal-2020-learning}, thus achieving increased performance on OOD data. 
Similar in spirit, \citet{wang-culotta-2020-identifying} have an expert annotating spurious \vs causal sentiment words and use word-level classification (spurious \vs genuine) to train robust classifiers that only rely on non-spurious words.
The cited works thus demonstrate that unwanted reliance on spurious patterns can be mitigated through extra annotation or (counterfactual) data generation. We further explore the latter option, and specifically focus on sentiment classification, as in \cite{kaushik2019learning,katakkar2021practical}.
Exploiting counterfactuals requires first to \begin{enumerate*}[(i)]
\item \label{it:cf-generate} generate them, and then
\item \label{it:cf-exploit} maximally benefit from  them in training.
\end{enumerate*}
For \ref{it:cf-exploit}, \citet{teney2020learning} present a loss term to leverage the relation between counterfactual and original samples.
In this paper we focus on \ref{it:cf-generate}, for which 
\citet{wu-etal-2021-polyjuice} use experts interacting with a finetuned GPT-2 \cite{radford2019language}. Alternatively, \citet{wang-etal-2021-enhancing, yang-etal-2021-exploring} use a pretrained language model and a sentiment lexicon.
Yet, having human annotators to create counterfactuals is still costly
(\eg 5~min/sample, \citet{kaushik2019learning}).
Thus, we pose the research question \textbf{(RQ)}: \emph{how to exploit a \textbf{limited} amount of counterfactuals to avoid classifiers relying on spurious patterns?}
We consider classifiers trained on representations obtained from frozen state-of-the-art sentence encoders \cite{reimers-gurevych-2019-sentence,gao-etal-2021-simcse}. 
We require only a few (human produced) counterfactuals, but artificially create additional ones based on them, directly in the encoding space (with a simple transformation of original instance representations), as sketched in \figref{fig:counterfactual-generation-idea}. This follows the idea of efficient sentence transformations 
in \citet{de-raedt-etal-2021-simple}.
\par We compare our approach against using
\begin{enumerate*}[(i)]
    \item \label{it:baseline:more-samples} 
    \emph{more} original samples and
    \item \label{it:baseline:counterfactuals}other models generating counterfactuals.
\end{enumerate*} 
We surpass both \ref{it:baseline:more-samples}--\ref{it:baseline:counterfactuals} for sentiment classification, with in-distribution and counterfactual training data from IMDb \cite{maas-etal-2011-learning,kaushik2019learning} and OOD-test data from Amazon \cite{ni-etal-2019-justifying}, SemEval \cite{rosenthal-etal-2017-semeval} and Yelp \cite{kaushik2020explaining}.
\section{Exploiting Few Counterfactuals}\label{sec:models}
\begin{table*}
\small
\centering
\begin{tabular}{p{3cm} p{5.8cm} p{5.8cm}}
\toprule
 & Original Sample ($x$) & Counterfactually Revised Sample ($\xc$) \\
\midrule
\textsc{negative} $\to$ \textsc{positive} & one of the \textcolor{negative}{\textbf{worst}} ever scenes in a sports movie. \textcolor{negative}{\textbf{3}} stars out of 10. & one of the \textcolor{positive}{\textbf{wildest}} ever scenes in a sports movie. \textcolor{positive}{\textbf{8}} stars out of 10.\\ \hline
\textsc{positive} $\to$ \textsc{negative} & The world of Atlantis, hidden beneath the earth’s core, is \textcolor{positive}{\textbf{fantastic}}. & The world of Atlantis, hidden beneath the earth’s core is \textcolor{negative}{\textbf{supposed to be fantastic}}. \\ 
\bottomrule
\end{tabular}
\caption{Two examples from \citet{kaushik2019learning} of counterfactual revisions made by humans for IMDb.}
\label{table:counterfactual_examples}
\end{table*}

We consider binary sentiment classification of input sentences/documents $x \in \X$, with associated labels $y \in \Y = \lbrace0, 1\rbrace$.
We denote the training set of labeled pairs $(x, y)$ as $\Did$, of size $n \triangleq \left| \Did  \right|$.
We further assume that for a limited subset of $k \ll n$ pairs $(x,y)$ we have corresponding manually constructed counterfactuals $(\xc, \yc)$, \ie $\xc$ is a minimally edited version of $x$ that has the opposite label $\yc = 1 - y$ (see \tabref{table:counterfactual_examples} for an example).
The resulting set of $k$ counterfactuals is denoted as $\Dcf$.
We will adopt a vector representation of the input $\phi(x)$, with $\phi: \X \to \Rd$.
We aim to obtain a classifier $f: \Rd \to \Y$ that, without degrading in-distribution performance, performs well on counterfactual samples and is robust under distribution shift.

\subsection{Exploiting Manual Counterfactuals}
\label{subsec:manual_augmented_models}

To learn the robust classifier $f$, we first present well-chosen reference approaches that leverage the $n$ in-distribution samples $\Did$ and the $k$ counterfactuals $\Dcf$. For all of the models below, we adopt logistic regression, but they differ in training data and/or loss function.

\noindent The \textbf{Paired} model
only uses the pairs for which we have counterfactuals, \ie the full set $\Dcf$ but only the corresponding $k$ pairs from $\Did$.

\noindent The \textbf{Weighted} model uses the full set of $n$ originals $\Did$, as well as all counterfactuals $\Dcf$, but compensates for the resulting data imbalance by scaling the loss function on $\Did$ by a factor $\frac{k}{n}$.
 
\subsection{Generating Counterfactuals}
\label{subsec:generated_augmented_models}
The basic proposition of our method is to artificially create counterfactuals for the $n-k$ original samples from $\Did$ that have no corresponding pair in $\Dcf$.
For this, we learn to map an original input document/sentence representation $\phi(x)$ to a counterfactual one, \ie a function $t: \Rd \to \Rd$.
We learn two such functions, $\tn$ to map representations of positive samples $\phi(x)$ (with $y=1$) to negative counterfactual representations $\phi(\xc)$ (with $\yc=0$), and vice versa for $\tp$.
We thus apply $\tn$ (respectively $\tp$) to the positive (resp.\ negative) input samples in $\Did$ for which we have no manually created counterfactuals.

\paragraph{Mean Offset}
Our first model is parameterless, where we simply add the average offset between representations of original positives $x$ (with $y=1$) and their corresponding $\xc$ to those positives for which we have no counterfactuals (and correspondingly for negatives).
Thus, mathematically:
$$\tn(\phi(x)) = \phi(x) + \on \, \mathrm{,with}$$
$$\on = \avg_{x\,:\,y = 1} \phi(\xc) - \phi(x)$$
(and correspondingly for $\tp$ based on counterfactuals of $x$ for which $y = 0$).

\paragraph{Mean Offset + Regression}
Since just taking the average offset may be too crude, especially as $k$ increases, we can apply an offset adjustment (noted as $r: \Rd \to \Rd$) learnt with linear regression.
Concretely, to create counterfactuals for positive originals we define:
$$\tn(\phi(x)) = \phi(x) + \on + \rn(\phi(x))$$ with a linear function 
$$\rn(\phi(x)) = \Wn \cdot \phi(x) + \bn$$
(learning $\Wn \in \Rdd$ and $\bn \in \Rd$ from the positive originals $x$ with corresponding counterfactuals $\xc$) and $\on$ as defined above. Similarly for $\tp$.

\section{Experimental Setup}
\label{sec:experiment}

\begin{table*}[h]
\small
\centering
\begin{tabular}{lcccccccc}
\toprule
 & \multicolumn{4}{c}{SimCSE-RoBERTa\textsubscript{\texttt{large}}} & \multicolumn{4}{c}{SRoBERTa\textsubscript{\texttt{large}}} \\
 \cmidrule(lr){2-5} \cmidrule(lr){6-9}
 \textit{Model} $\ $ 
 ($n$)\,($k$)
 & Orig. (\%) & CAD (\%) & OOD (\%) & Avg. & Orig. (\%) & CAD (\%) & OOD (\%) & Avg.   \\ 
\midrule
Original \ (3.4k)\,(0)  & 89.6$_{\pm0.7}$ & 75.7$_{\pm1.2}$ & 74.6$_{\pm2.6}$ & 80.0 & 90.7$_{\pm0.6}$ & 78.8$_{\pm1.7}$ & 80.6$_{\pm2.4}$ & 83.4 \\ 
\midrule
Weighted \ (1.7k)\,(16) & 88.1$_{\pm0.8}$ & 78.5$_{\pm1.1}$ & 75.1$_{\pm2.3}$ & 80.6 &  89.2$_{\pm0.8}$ & 81.1$_{\pm1.3}$ & 82.9$_{\pm2.1}$ & 84.4 \\
Paired \ (16)\,(16)  & 81.5$_{\pm2.2}$ & 80.9$_{\pm2.4}$ & 77.5$_{\pm4.3}$ & 80.0 & 86.9$_{\pm1.3}$ & 77.9$_{\pm2.2}$ & 83.9$_{\pm4.2}$ & 82.9 \\
\midrule
\multicolumn{9}{l}{\textbf{\citet{wang2021robustness}:} \ (1.7k)\,(0)} \\
- Pred.\ from top \ ($n''$=1,284) &
81.4 & 82.6 & 73.0 & 79.0 & 83.6 & 83.4 & 73.4 & 80.1 \\
- Ann.\ from top \ ($n''$=1,618)&
80.3 & 84.2 & 74.1 & 79.5 & 81.8 & 86.1 & 71.2 & 79.7 \\
- Ann.\ from all \ ($n''$=1,694)
& 83.0 & 85.7 & 76.5 & 81.7 & 85.7 & 89.8 & 75.6 & 83.7 \\
\midrule
\multicolumn{9}{l}{\textbf{Our models:} \ (1.7k)\,(16)} \\
- Mean Offset  &  86.2$_{\pm1.2}$ & 84.6$_{\pm1.3}$ & 78.0$_{\pm3.2}$ & \textbf{83.0} &  88.1$_{\pm1.2}$ & 85.6$_{\pm1.1}$ & 83.0$_{\pm3.3}$ & \textbf{85.6} \\
- Mean Offset + Regression 
& 86.1$_{\pm1.2 }$& 84.1$_{\pm1.3}$ & 78.2$_{\pm3.1}$ & 82.8 & 88.3$_{\pm1.0}$ & 85.2$_{\pm1.5}$ & 83.4$_{\pm3.3}$ & \textbf{85.6} \\
\bottomrule
\end{tabular}
\caption{
Results with $k$ = 16 manually crafted counterfactuals and $n$ original samples. Note that our models then use an additional $n' = n - k$ artificial counterfactuals generated in representation space. The models of \citet{wang2021robustness} automatically generate $n''$ counterfactuals in the input space.
}
\label{table:main_table}
\end{table*}

\paragraph{Datasets} For the in-distribution data, we use a training set $\Dset{train}{id}$ of 1,707 samples, and a test set $\Dset{test}{id}$ of 488 samples, with all of these instances randomly sampled from the  original IMDb sentiment dataset of 25k reviews \cite{maas-etal-2011-learning}. 
The counterfactual sets $\Dset{train}{cad}$ and $\Dset{test}{cad}$ 
are the revised versions of $\Dset{train}{id}$ and $\Dset{test}{id}$, as rewritten by Mechanical Turk workers recruited by \citet{kaushik2019learning}. 
See \appref{app:implementation_details} for  further details.
We will also test on out-of-distribution (OOD) data from Amazon \cite{ni-etal-2019-justifying}, SemEval \cite{rosenthal-etal-2017-semeval} and Yelp \cite{kaushik2020explaining} (we note these datasets as $\Damzn$, $\Dsemeval$, $\Dyelp$).

\paragraph{Sentence Encoders} To obtain $\phi(x)$, we use the sentence encoding frameworks SBERT and SimCSE (\citet{reimers-gurevych-2019-sentence, gao-etal-2021-simcse}). 
The main results are presented with SRoBERTa\textsubscript{\texttt{large}} and SimCSE-RoBERTa\textsubscript{\texttt{large}} and they are kept frozen at all times. \appref{app:implementation_details} lists additional details; \appref{app:additional_results} shows results for other encoders.
%
\paragraph{Baselines}
As a baseline for our few-counterfactuals-based approaches, we present results (\emph{Original}) from a classifier trained on twice the amount of original (unrevised, in-distribution) samples. Further, we also investigate competitive counterfactual-based approaches as proposed by \citet{wang2021robustness}, who leverage identified causal sentiment words and a sentiment lexicon to generate counterfactuals in the input space
(which we subsequently embed with the same sentence encoders $\phi$ as before). They adopt three settings, with increasing human supervision, to identify causal words:
\begin{enumerate*}[(i)]
    \item \emph{predicted from top}: 32 causal words were identified automatically for IMDb;
    \item \emph{annotated from top}: a human manually marked 65 words as causal from a top-231 word list deemed most relevant for sentiment; and
    \item \emph{annotated from all}: a human labeled 282 causal words from the full 2,388 word vocabulary.
\end{enumerate*}

\paragraph{Training and Evaluation}
 For all presented approaches, the classifier $f$ is implemented by 
 logistic regression with L2 regularization, where the regularization parameter $\lambda$ is established by 4-fold cross-validation.\footnote{We experiment with both weak and strong regularization. See \appref{subsec:impact_regularization} for details.}
The results presented further in the main paper body report are obtained by training on the complete training set (\ie all folds).

 The \emph{Mean Offset + Regression} model of \secref{subsec:generated_augmented_models}, to artificially generate counterfactuals, is implemented by linear regression with ordinary least squares.
 The \emph{Weighted} and \emph{Paired} classifiers of \secref{subsec:manual_augmented_models} are trained on $n$ samples from $\Dset{train}{id}$ together with $k$ counterfactuals sampled from $\Dset{train}{cad}$.
 To evaluate our classifiers with \emph{generated} counterfactuals, as described in  \secref{subsec:generated_augmented_models}, we train
 on the $n$ original samples, $k$ manual counterfactuals and $n-k$ generated counterfactuals.
 The \emph{Original} baseline uses $2 \cdot n$ original samples $\Dset{train}{id} \cup \Dset{train}{id}\,'$, adding an extra $|\Dset{train}{id}\,'| = n$ that are sampled randomly from the 25k original, unrevised IMDb reviews (but not in $\Dset{train}{id}$ and $\Dset{test}{id}$).
For the counterfactual-based models of \citet{wang2021robustness}, the training set is expanded with $n'' \approx n$ counterfactuals (based on $\Dset{train}{id}$) automatically generated, in the input space.

 We evaluate the accuracy on $\Dset{test}{id}$, $\Dset{test}{cad}$ and the OOD test sets $\Damzn$, $\Dsemeval$, $\Dyelp$ (averaging the accuracies over these 3 sets for OOD evaluation).
 For each $k \in \lbrace16,32,\ldots,128\rbrace$, we use 50 different random seeds to sample:
 \begin{enumerate*}[(i)]
     \item $k/2$ negative and $k/2$ positive counterfactuals, and
     \item $n$ additional original samples (for \emph{Original} baseline).
 \end{enumerate*}
The reported accuracies are averaged across the 50 seeds.
\section{Results and Discussion}
\label{sec:results}

\begin{figure*}[t]
    \centering
    \begin{subfigure}{0.25\textwidth}
      \includegraphics[height=3.5cm,width=\textwidth]{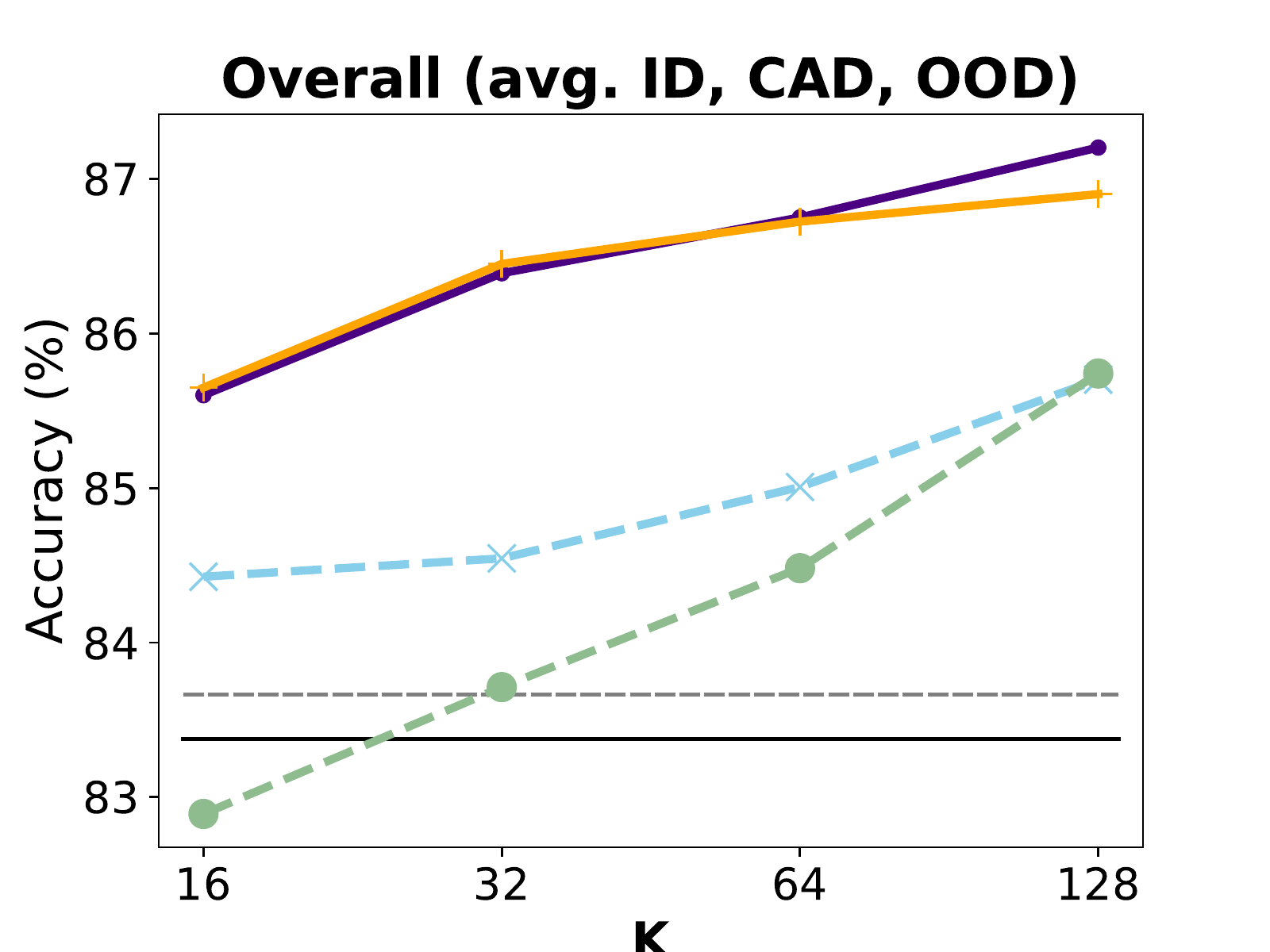}
    \end{subfigure}\hspace{-1em}
    \begin{subfigure}{0.25\textwidth}
      \includegraphics[height=3.5cm, width=\textwidth]{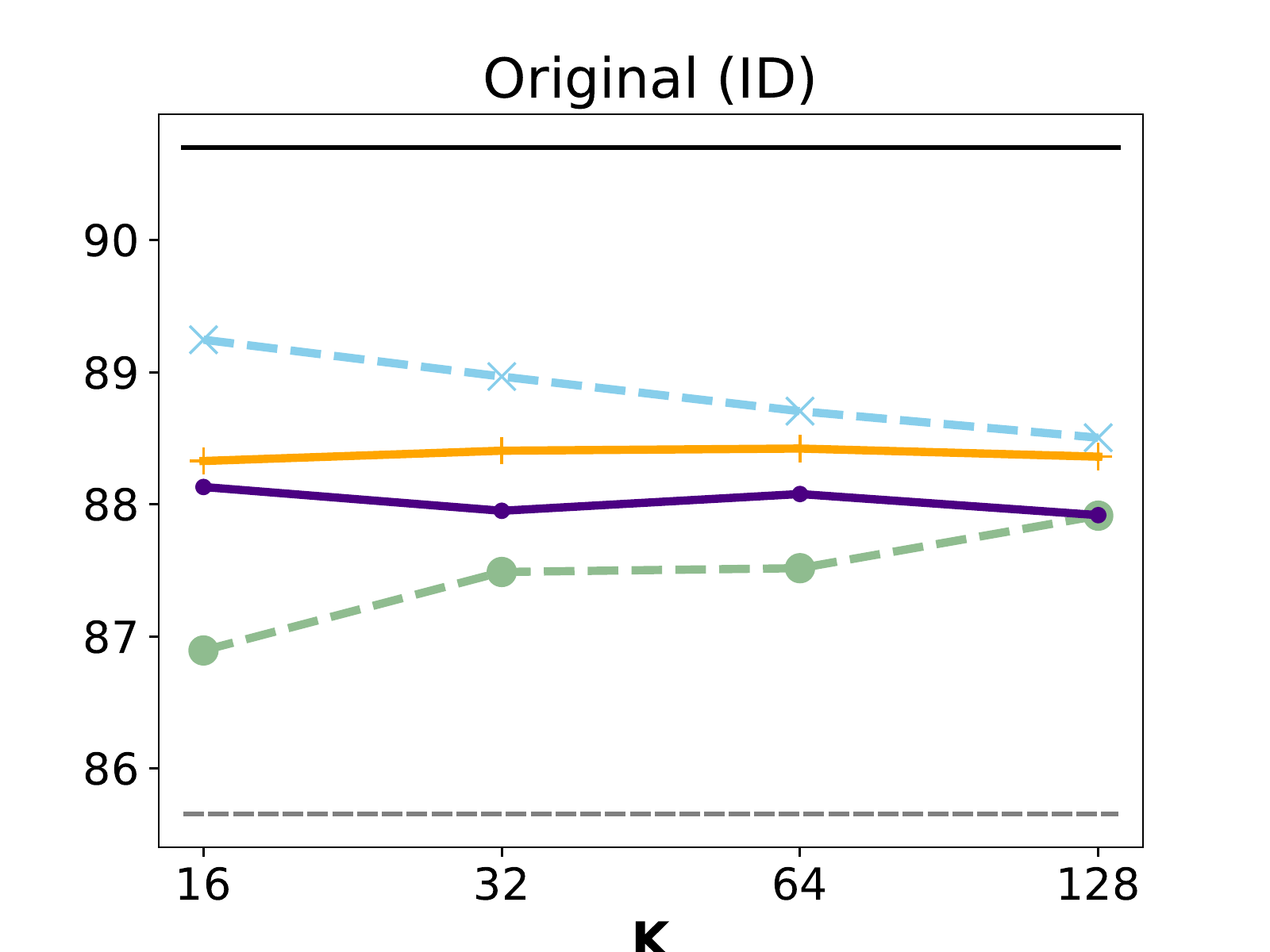}
    \end{subfigure}\hspace{-1em}
    \begin{subfigure}{0.25\textwidth}
      \includegraphics[height=3.5cm,width=\textwidth]{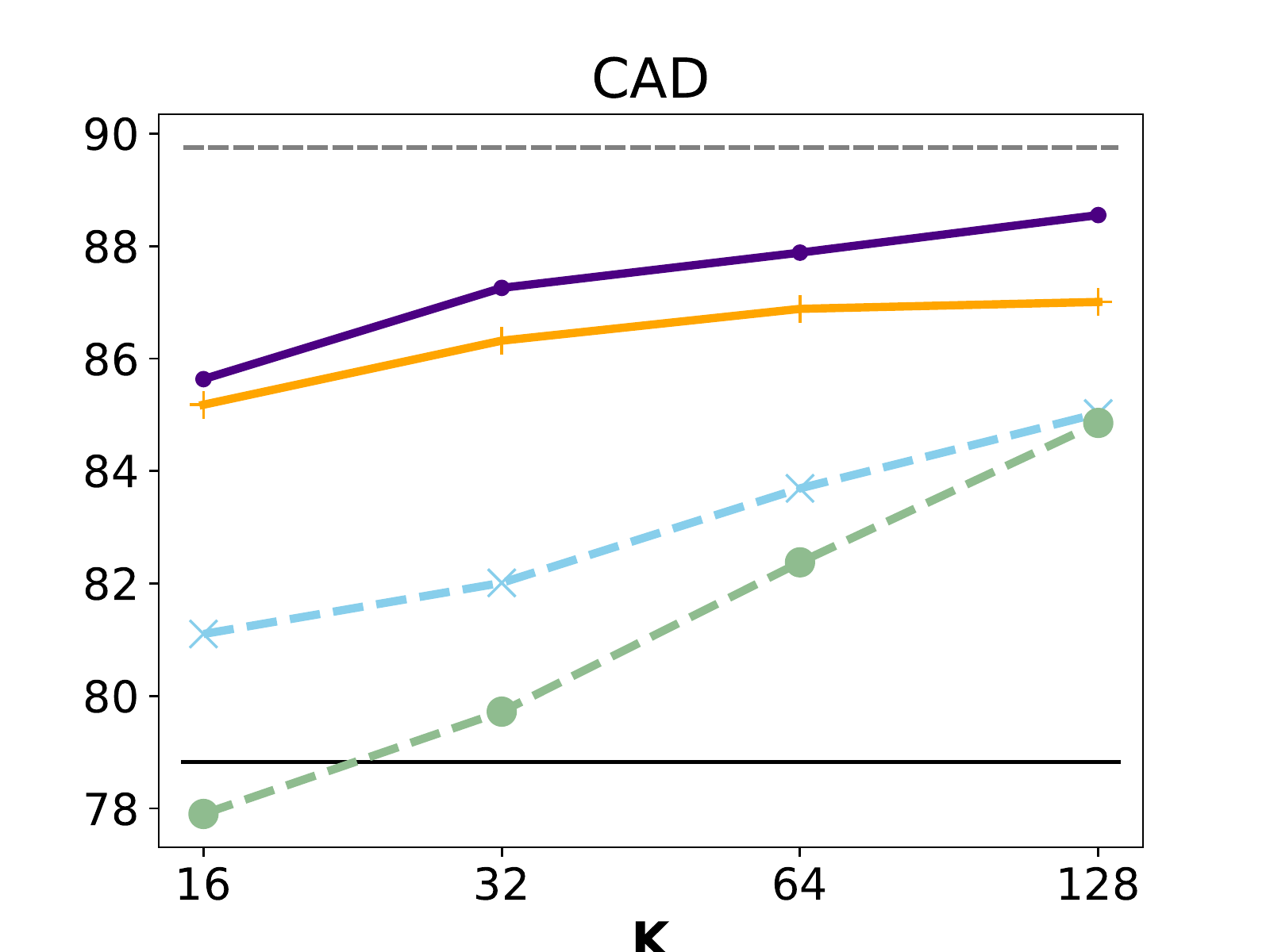}
    \end{subfigure}\hspace{-1em}
    \begin{subfigure}{0.25\textwidth}
      \includegraphics[height=3.5cm,width=\textwidth]{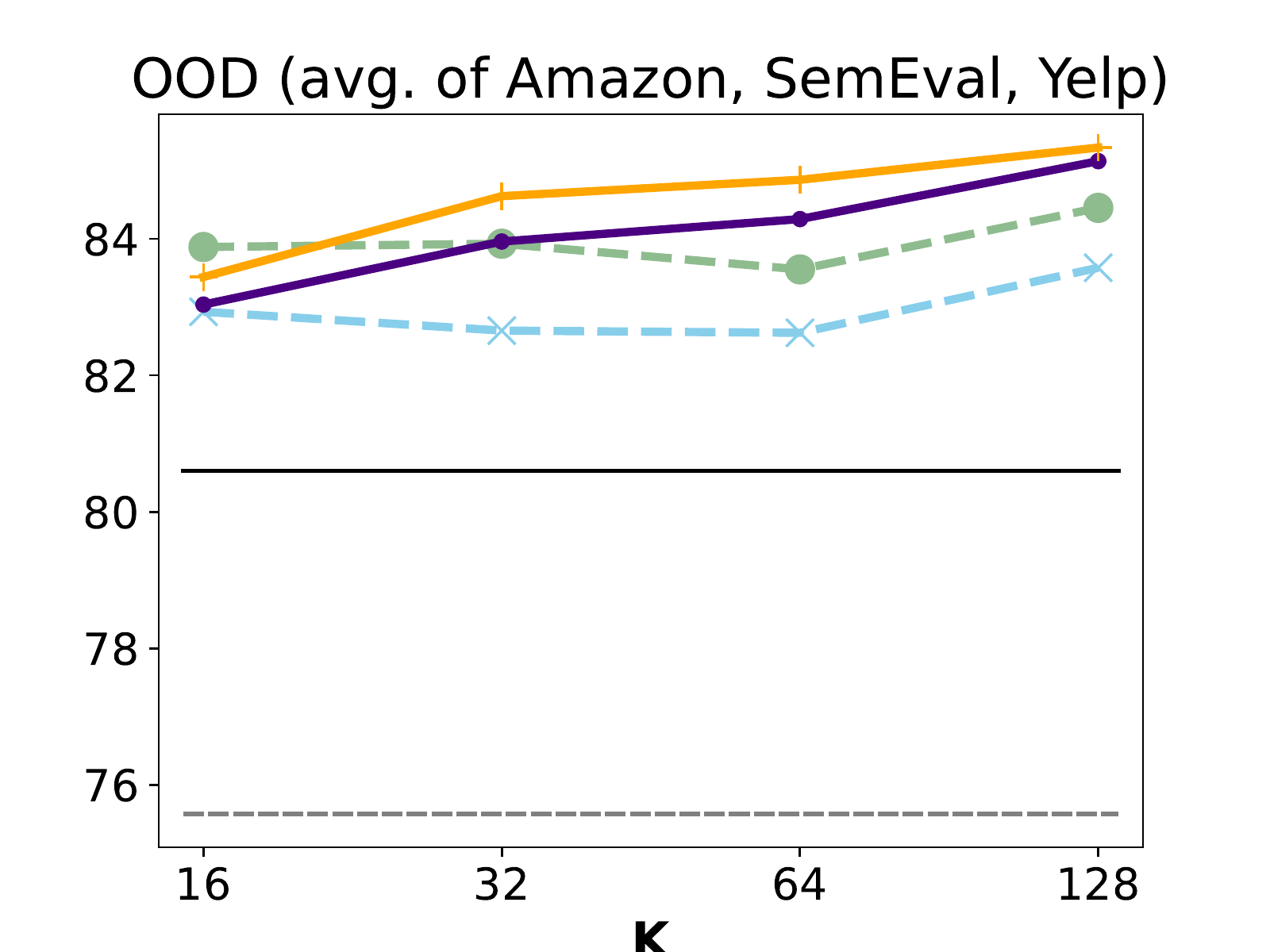}
    \end{subfigure}
    \begin{subfigure}{1.\textwidth}
      \includegraphics[width=\textwidth]{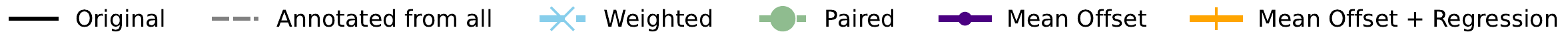}
      \label{fig:legend}
    \end{subfigure}\vspace{-1.5em}
    \caption{Accuracies for an increasing number of manually crafted counterfactuals $k$ for SRoBERTa\textsubscript{\texttt{large}}-based  models. From  \citet{wang2021robustness}, we selected the best performing model (Annotated from all).} 
    \label{fig:main_figure}
\end{figure*}

\paragraph{Main Results} We investigate if a \emph{limited} amount of counterfactuals suffices to make classifiers less sensitive to spurious patterns: 
classification should also perform well on counterfactual and OOD examples, without sacrificing in-distribution (ID) performance.

\Figref{fig:main_figure} and  \tabref{table:main_table} report on the accuracies for the different models on the original ($\Dset{test}{id}$), counterfactual ($\Dset{test}{cad}$) and OOD test sets (avg.\ over $\Damzn$, $\Dsemeval$, $\Dyelp$). 
The average of these three values is plotted as the \emph{overall} metric in the leftmost panel of \figref{fig:main_figure} and the Avg.\ column in \tabref{table:main_table}.
From this overall perspective, we note that our classifiers trained on offset-based counterfactuals outperform the \emph{Original} baseline trained on 3.4k samples by +3\% (+2.2\%) points in accuracy for SimCSE-RoBERTa\textsubscript{\texttt{large}} (SRoBERTa\textsubscript{\texttt{large}}),
even when the number of manually crafted counterfactuals is less than 1\% (\ie for $k$\,=\,16) of the original sample size ($n$\,=\,1.7k).
Moreover, we note that all counterfactual-based classifiers improve for increasing numbers of counterfactuals $k$. We observe little difference between the performance of classifiers trained on generated counterfactuals from the \emph{Mean Offset} and the \emph{Mean Offset\,+\,Regression} models, with the former working slightly better than the latter for larger $k$, indicating that the simple mean offset is a good choice. 
Moreover, our classifiers trained on offset-based counterfactuals of SimCSE-RoBERTa\textsubscript{\texttt{large}} (SRoBERTa\textsubscript{\texttt{large}}) show a clear improvement over both classifiers \begin{enumerate*}[(i)]
    \item without generated counterfactuals (+2.4\% (+1.2\%) over \emph{Weighted} and +3\% (+2.7\%) over \emph{Paired}), and
    \item trained on counterfactuals generated with the best model of \citet{wang2021robustness} (improving by +1.3\% (+1.9\%)).
\end{enumerate*}
\noindent (ii) relies on annotating a 2,388 word vocabulary (which we speculate to be more labor-intensive than creating just 16 counterfactuals; poor \emph{Predicted (Annotated) from top} results suggest we cannot avoid human annotation).

We further compare the models to the \emph{Original} baseline on the three test subsets (ID, CAD, OOD).
For straightforward imbalance counteracting strategies (\emph{Paired}, \emph{Weighted}), we observe expected performance improvement of SimCSE-RoBERTa\textsubscript{\texttt{large}} (SRoBERTa\textsubscript{\texttt{large}}) for data that deviates from the ID training data, \ie for OOD and CAD --- yet, clearly more advanced methods like ours do way better --- while sacrificing performance for ID itself.
We find that the best \citet{wang2021robustness} model excels at CAD with improvements of +10\% (+11\%), but by doing so suffers a lot on ID $-$6.6\% ($-$5\%) and on OOD +1.9\% ($-$5\%). 
In contrast, our \emph{Mean Offset} model strikes the desirable balance across ID, CAD, and OOD performance with a smaller drop in ID accuracy of $-$3.4\% ($-$2.6\%), and with improvements on both CAD and OOD of respectively +8.9\% (+6.8\%) and +3.4\% (+2.4\%).

\paragraph{Ablations}\label{sec:ablation}
\begin{table*}
\small
\centering
\addtolength{\tabcolsep}{-0.5pt}
\begin{tabular}{lcccccccc}
\toprule
 & \multicolumn{4}{c}{SimCSE-RoBERTa\textsubscript{\texttt{large}}} & \multicolumn{4}{c}{SRoBERTa\textsubscript{\texttt{large}}} \\
 \cmidrule(lr){2-5} \cmidrule(lr){6-9}
 \textit{Model} $\ $ ($n$)\,($k$)
 & Orig. (\%) & CAD (\%) & OOD (\%) & Avg. & Orig. (\%) & CAD (\%) & OOD (\%) & Avg.   \\ 
\midrule
\multicolumn{9}{l}{\textbf{Ablation models}:}  \\
- Random Offset (1.7k)\,(0) & 87.7$_{\pm0.7}$ & 74.3$_{\pm1.0}$ & 73.1$_{\pm2.6}$ & 78.3 & 88.6$_{\pm1.0}$ & 77.9$_{\pm1.3}$ & 79.0$_{\pm3.3}$ & 81.8\\
- Mean\textsubscript{\textsc{id}} Offset (1.7k)\,(0) & 88.9$_{\pm0.3}$ & 76.1$_{\pm0.2}$ & 73.8$_{\pm0.2}$ & 79.6 & 88.4$_{\pm0.3}$ & 79.0$_{\pm0.2}$ & 78.9$_{\pm0.8}$ & 82.1 \\
- Linear Regression (1.7k)\,(16) & 88.2$_{\pm0.9}$ & 77.8$_{\pm1.5}$ & 74.9$_{\pm2.4}$ & 80.3 & 89.5$_{\pm0.8}$ & 81.8$_{\pm1.0}$ & 83.2$_{\pm1.7}$ & 84.8 \\
\midrule 
\multicolumn{9}{l}{\textbf{Our models}: \ (1.7k)\,(16)} \\
- Mean Offset  &  86.2$_{\pm1.2}$ & 84.6$_{\pm1.3}$ & 78.0$_{\pm3.2}$ & \textbf{83.0} &  88.1$_{\pm1.2}$ & 85.6$_{\pm1.1}$ & 83.0$_{\pm3.3}$ & \textbf{85.6} \\
- Mean Offset + Regression  & 86.1$_{\pm1.2 }$& 84.1$_{\pm1.3}$ & 78.2$_{\pm3.1}$ & 82.8 & 88.3$_{\pm1.0}$ & 85.2$_{\pm1.5}$ & 83.4$_{\pm3.3}$ & \textbf{85.6} \\
\bottomrule
\end{tabular}
\addtolength{\tabcolsep}{0.5pt}
\caption{\textbf{Ablations} ($k$\,=\,16): A comparison of our models with models that generate counterfactuals by (i) adding a random offset (with same L2-norm as the mean offset) to original samples, (ii) adding the mean offset calculated between the $n$ original samples with opposite labels (without $k$ counterfactuals) to original samples or (iii) by transforming the original samples directly with linear regression (\ie without the mean offset).
}\label{table:ablation}
\end{table*}
To investigate the effectiveness of our \emph{Mean Offset (+ Regression)} approaches that exploit $k$ manually crafted counterfactuals, we provide ablations by including the scores for counterfactuals generated with
\begin{enumerate*}[(i)]
    \item a \emph{Random Offset} with the same L2-norm as the mean offset,
    \item a \emph{Mean\textsubscript{\textsc{id}} Offset} calculated among the $n$ original samples with opposite labels (and thus without $k$ manual counterfactuals), and
    \item a mapping function $t$ modeled directly 
    with \emph{Linear Regression}, \ie $t(\phi(x)) = W\cdot\phi(x) + b$ (with $b \in \mathbb{R}$, $W \in \mathbb{R}^{d \times d}$).
\end{enumerate*}

Following \secref{sec:experiment}, we randomly sample 50 times: random offsets, and $k$ (original, counterfactual) pairs from $\Dset{train}{id}$ and $\Dset{train}{cad}$ from which the \emph{Mean Offset} is calculated and from which the parameters of \emph{Mean Offset + Regression} are learnt.

The classification accuracies for the ablation models are shown in \tabref{table:ablation}, demonstrating the importance of using counterfactuals to calculate an effective offset: the \emph{Mean Offset} consistently outperforms the \emph{Random Offset} and the \emph{Mean\textsubscript{\textsc{id}} Offset} (both calculated without manual counterfactuals). Since the \emph{Random Offset} shifts in a totally arbitrary direction, it does not produce very useful ``counterfactuals'' to learn from. As the \emph{Mean\textsubscript{\textsc{id}} Offset} is calculated among the $n$ original samples, it does not provide ``new'' information that was not already present in the original samples. 

At last, we observe expected performance improvement for the \emph{Mean Offset (+ Regression)} over the \emph{Linear Regression} model since directly learning its transformation matrix ($W \in \mathbb{R}^{d \times d}$) from just $k$ (=16) (original, counterfactual) pairs is difficult.

\section{Conclusion}
We explored improving the robustness of classifiers (\ie make them perform well also on out-of-distribution, OOD, data) by relying on just a few manually constructed counterfactuals.
We propose a simple strategy, learning from  few (original, counterfactual) pairs how to transform originals $\Did$ into counterfactuals $\Dcf$, in a document vector representation space: shift an original document with the mean offset among the given pairs.
Thus, using just a small number (1\% of $|\Did|$) of manual counterfactuals, we outperform sentiment classifiers trained using either
\begin{enumerate*}[(i)]
    \item 100\% extra original samples, or
    \item a state-of-the-art (lexicon-based) counterfactual generation approach.
\end{enumerate*}
Thus, we suggest that additional annotation budget is better spent on counterfactually revising available annotations, rather than collecting similarly distributed new samples.

\section*{Acknowledgements}
This work was funded by the Flemish Government (VLAIO), Baekeland project-HBC.2019.2221.

\section*{Limitations}
Our work is limited in terms of \emph{interpretability}, the \emph{trade-off between computational efficiency and model effectiveness}, and the \emph{application domain} of the presented experiments. These limitations are discussed in the following paragraphs. 
\paragraph{Interpretability}
 Our models produce counterfactual samples directly in the \emph{encoding vector space}, and they thus cannot easily be interpreted. One could train a decoder to reconstruct the IMDb documents from the frozen vector representations. However, we believe it to be infeasible given \begin{enumerate*}[(i)]
    \item the considerable length of the IMDB documents (more than 160 words on average), and
    \item the fact that the document would need to be decoded from a single vector without relying, e.g., on the attention mechanism (since the decoder should otherwise bypass the single vector representation)
\end{enumerate*}. Thus, it would be hard to discern whether observed noise in the reconstructed
full review documents is due to flaws in our generated vectors, or rather the imperfect
decoder. Hence, we opted for the quantitative analysis in \appref{app:analysis_generated_counterfactuals} instead.

\paragraph{Efficiency vs.~effectivenes}
Our methods generate counterfactual vectors in the encoding space of \emph{frozen} sentence encoders such that the attained accuracy, while competitive, may be lower than when compared to fully fine-tuned transformers. However, leveraging frozen sentence encoders allows us to
train way faster (<1 minute on CPU): the linear sentiment classification layer contains less than 2K parameters, estimating the mean offset is parameterless, and the linear transformation of the mean offset + regression model contains less than 1.2M parameters. In contrast, fully finetuning BERT requires updating all 110M parameters for 20 epochs on a Tesla V100 GPU \cite{kaushik2019learning}. In addition, using pre-trained \emph{frozen} sentence encoders allows us to analyze whether their produced embeddings are able to model the subtle differences between original and counterfactual samples, and whether this difference can be exploited to improve robustness.

\paragraph{Application domain}
We presented results for sentiment classification, given that, to the best of our knowledge, the only \emph{topic} classification datasets with paired counterfactual training samples is IMDb \cite{kaushik2019learning}.  However, we believe that our method could generalize beyond sentiment classification to other \emph{topic} classification tasks for which there is a clear direction in the vector space between different topics (=classes), such as from positive to negative (or vice versa). Note that this is not the case for the Natural Language Inference (NLI) task. Hence, why we did not experiment on existing NLI datasets with available counterfactuals \cite{kaushik2019learning}.

\bibliography{anthology,custom}

\clearpage 

\appendix
\label{sec:appendix}
\section*{Appendices}
\section{Additional Results}
\label{app:additional_results}

We provide additional results for other sentence encoders
\begin{enumerate*}[(i)]
    \item for an increasing number of counterfactuals $k$ in Figs.\ \ref{fig:sroberta-large}--\ref{fig:smpnet} and
    \item for $k=16$ in Tables \ref{table:sroberta-large}--\ref{table:simcse-bert-base}.
\end{enumerate*}
We discuss the impact of regularization in \secref{subsec:impact_regularization}, provide an analysis of the generated counterfactuals in \secref{app:analysis_generated_counterfactuals}, discuss the overall robustness in \secref{subsec:overal_robustness}, and compare the strengths and weaknesses of the different approaches in \secref{subsec:individual_robustness}.

\subsection{Impact of Regularization Strength}
\label{subsec:impact_regularization}
The methods using generated counterfactuals expand the original training set $\Dset{train}{id}$ with counterfactuals $\Dset{train}{cad}$. Our models add $k$ manually crafted ones and $n-k$ generated ones, whereas the state-of-the-art models \cite{wang2021robustness} add $n''\approx n$ generated ones.
When using both ID and CAD samples for training, we risk overfitting to the generated counterfactuals (which may be more narrowly distributed than data in the wild, for which the OOD and CAD test samples are a proxy).
Such overfitting can be avoided by enforcing stronger L2 regularization (\ie larger $\lambda$). 
We analyze whether this is useful and experiment by either
\begin{enumerate*}[(i)]
    \item \label{it:reg-free} \emph{free regularization}, allowing\footnote{Recall that we pick the best $\lambda$ through cross-validation, see \secref{sec:experiment}.} a broad range spanning both weak and strong regularization $\lambda \in \lbrace 10^{-3},10^{-2},\ldots, 10^3\rbrace$, or
    \item \label{it:reg-strong} \emph{strong regularization} restricting the choice to $\lambda \in \lbrace 1, 10,\ldots, 10^3\rbrace$.
\end{enumerate*} 
Tables~\ref{table:sroberta-large}--\ref{table:simcse-bert-base} show the results for such free \vs strong regularization.
We note that both model types using artificially generated counterfactuals (ours and \citet{wang2021robustness}) generally
benefit from \emph{strong} regularization, whereas the others perform better under \emph{free} regularization (suggesting they are less prone to overfitting).
We note that especially models  based on \citet{wang2021robustness} may suffer from overfitting (\eg a difference of almost 14 percentage points on OOD accuracy even for their best model, \emph{Ann.\ from all}, based on SimCSE-RoBERTa\textsubscript{\texttt{large}}).
In conclusion, as main paper results we therefore reported the results of the best \emph{free} regularization $\lambda$ choice for models without generated counterfactuals, whereas for ours and \citet{wang2021robustness} we reported \emph{strong} regularization results.

\subsection{Analysis of Generated Counterfactual Vectors}\label{app:analysis_generated_counterfactuals}

\begin{table*}
\small
\centering
\begin{tabular}{lcccccc}
\toprule
& \multicolumn{3}{c}{SimCSE-RoBERTa\textsubscript{\texttt{large}}} &
\multicolumn{3}{c}{SRoBERTa\textsubscript{\texttt{large}}} 
 \\ \cmidrule(lr){2-4} \cmidrule(lr){5-7}
\textit{Samples}  & R$^2$ & RMSE & Diversity & R$^2$ & RMSE & Diversity \\
\midrule
Original samples ($\Dset{test}{id}$)
& \phantom{$-$}\textbf{0.747}\phantom{$\,\downarrow$} & \num{0.0001322} & 0.549\phantom{$\,\downarrow$}
& \textbf{0.797} & \num{0.0001214} & 0.627\phantom{$\,\downarrow$}\\
Manual Counterfactuals ($\Dset{test}{cad}$)
& - & - & \textbf{0.539}\phantom{$\,\downarrow$} 
& - & - & \textbf{0.621}\phantom{$\,\downarrow$}\\
 \midrule 
 \multicolumn{7}{l}{\textbf{Generated from ablation models}:}  \\
- Linear Regression  
& $-$0.066\,$\downarrow$ & \num{0.0232278} & 0.087\,$\downarrow$
& $-$0.068$\,\downarrow$ & \num{0.0251788} & 0.092$\,\downarrow$ \\
- Random Offset  
& \phantom{$-$}0.654$\,\downarrow$ & \num{0.0126325} & 0.524\,$\downarrow$ 
& \phantom{$-$}0.724\,$\downarrow$ & \num{0.0121939} & 0.601$\,\downarrow$ \\
\midrule 
\multicolumn{7}{l}{\textbf{Generated from our models}:}  \\ 
- Mean Offset  
& \phantom{$+$}0.785\,$\uparrow$ & \num{0.0103255} & 0.537$\,\approx$ 
& \phantom{$-$}0.830\,$\uparrow$ & \num{0.0099161} & 0.619$\,\approx$ \\
- Mean Offset + Regression
& \phantom{$+$}0.779$\,\uparrow$ & \num{0.0104737} & 0.536$\,\approx$  
& \phantom{$+$}0.821$\,\uparrow$ & \num{0.0101802} & 0.618$\,\approx$\\
\bottomrule
\end{tabular}
\caption{\textbf{Analysis of generated counterfactuals} ($k$\,=\,16): R$^2$ and RMSE scores between the encoded samples of $\Dset{test}{ID}$ and  $\Dset{test}{cad}$ as a reference (in \textbf{bold}), and between the generated counterfactuals and encodings in $\Dset{test}{cad}$, where $\uparrow$ and $\downarrow$ denote values higher or lower than the reference.
We compare diversity of the generated counterfactuals compared to that of the encodings of manually crafted ones ($\Dset{test}{cad}$) (in \textbf{bold}), where $\approx$ and $\downarrow$ respectively denote similar and lower diversity.}
\label{table:analysis}
\end{table*}

Since our methods produce counterfactuals in the encoding space, they
cannot be easily interpreted.
Still, we attempt to analyze how well they are aligned with manual constructed counterfactuals.
To do so we measure
\begin{enumerate*}[(i)]
    \item the coefficient of determination, R$^2$, and
    \item the root mean squared error (RMSE) between the generated and manual counterfactual \emph{test vectors}.
\end{enumerate*}
Both metrics quantify how well the generated vectors approximate manual counterfactual vectors.
In addition, we provide a measure of diversity, calculated as the average pairwise \emph{cosine distance} among generated samples: we compare it against that diversity among vectors of the manually constructed counterfactuals. A well approximated set of generated counterfactuals should be as diverse as a manually constructed set, and approach the unrevised originals' diversity.\footnote{Observe that the diversity of counterfactuals $\Dset{test}{cad}$ is slightly lower than the corresponding originals $\Dset{test}{id}$, suggesting that edits are less diverse than the original phrasings.}

\paragraph{Setup} Following the same setup as in \secref{sec:experiment}, we randomly sample 50 times: $k$ (original, counterfactual) pairs from $\Dset{train}{id}$ and $\Dset{train}{cad}$, from which  the \emph{Mean Offset} is calculated and from which the parameters of the \emph{Mean Offset + Regression} are learned. We apply both transformations on the original test \emph{encodings} ($\Dset{test}{id}$) to generate counterfactuals and compare them to the encodings of the manual test samples ($\Dset{test}{cad}$). 
\par Moreover, we provide an ablation by including the scores for counterfactuals generated with two ablation models of \secref{sec:ablation}, \ie
\begin{enumerate*}[(i)]
\item a \emph{Random Offset} (with same L2-norm as the mean offset),
\item a mapping function $t$ modeled directly\footnote{I.e., rather than through a linear regressor for the residual offset $r(\phi(x))$ relative to the mean  $\Vec{o}$ as defined in \secref{sec:models}.} \emph{Linear Regression}, \ie $t(\phi(x)) = W\cdot\phi(x) + b$ (with $b \in \mathbb{R}$, $W \in \mathbb{R}^{d \times d}$).
\end{enumerate*}   
\par We also report the R$^2$ and RMSE scores between the original encodings and their manually revised counterparts, which we use as a reference to 
determine to what extent generated counterfactual encodings align with those of manually crafted counterfactuals.

\paragraph{Results} The results are shown in \tabref{table:analysis} (averaged over the 50 runs).
First, we assess how well the generated counterfactuals approximate the manual ones using R$^2$ scores.
We observe that R$^2$ between the original samples and their corresponding manual counterfactual already reaches 0.747 and 0.797 for SimCSE-RoBERTa\textsubscript{\texttt{large}} and SRoBERTa\textsubscript{\texttt{large}} respectively.
This is not surprising, since only a minimal number of words are edited in revising an original sample to its counterfactual.
The generated counterfactuals improve over the original score, with R$^2$ scores for the Mean Offset (+ Regression) of 0.785 (0.779) and 0.830 (0.821), respectively for SimCSE-RoBERTa\textsubscript{\texttt{large}} and SRoBERTa\textsubscript{\texttt{large}}.
Conversely, the ablation models result in counterfactuals more dissimilar than the originals (even lower R$^2$ than our models), with especially the \emph{Linear Regression} model performing poorly.
Additionally, the RMSE-scores for our models are notably lower than those from the ablation models, but
clearly larger than the (very low, because of minimal edits) RMSE for manual counterfactuals.

Second, we assess that the generated counterfactuals preserve the original diversity for both encoders, with scores for the Mean Offset (+~Regression) method of 0.537 (0.536) and 0.619 (0.618) that are very close to those for the manual counterfactuals (0.539 and 0.621, respectively for the two encoders). 
The ablation models on the other hand attain lower diversity scores, with especially \emph{Linear regression} behaving extremely poorly ---
which we suspect to be caused by a collapse where the same subset of vectors are predicted regardless the input.

Third, when looking at both \tabref{table:ablation} (classification accuracies for the ablation models \secref{sec:ablation}) and \tabref{table:analysis}, we observe that the \emph{Linear Regression} model outperforms \emph{Random Offset} overall in terms of attained accuracies (given its better performance for CAD and OOD),
even though \emph{Random Offset}'s ``better'' counterfactuals (\cf higher R$^2$ and lower RMSE in \tabref{table:analysis}) may lead one to expect the opposite. We speculate that, given the very low diversity, the \emph{Linear Regression} model just predicts the set of $k$ counterfactuals which it saw during training, making it more similar to the \emph{Weighted} model that does not train with generated counterfactuals.

\subsection{Overall Robustness}\label{subsec:overal_robustness} 
\begin{table*}
\small
\centering
\begin{tabular}{lcccccccc}
\toprule
 & \multicolumn{4}{c}{SimCSE-RoBERTa\textsubscript{\texttt{large}}} & \multicolumn{4}{c}{SRoBERTa\textsubscript{\texttt{large}}} \\
 \cmidrule(lr){2-5} \cmidrule(lr){6-9}
 \textit{Model} $\ $ ($n$)\,($k$)
 & Orig. (\%) & CAD (\%) & OOD (\%) & Avg. & Orig. (\%) & CAD (\%) & OOD (\%) & Avg.   \\ 
\midrule
- Original (3.4k)\,(0)  & 89.6$_{\pm0.7}$ & 75.7$_{\pm1.2}$ & 74.6$_{\pm2.6}$ & 80.0 & 90.7$_{\pm0.6}$ & 78.8$_{\pm1.7}$ & 80.6$_{\pm2.4}$ & 83.4 \\ 
- Original (24k)\,(0) & 91.1$_{\pm0.0}$ & 78.5$_{\pm0.0}$ & 77.8$_{\pm0.0}$ & 82.5 & 92.6$_{\pm0.0}$ & 80.1$_{\pm0.0}$ & 80.9$_{\pm0.0}$ & 84.6 \\
\midrule 
\multicolumn{9}{l}{\textbf{Our Models}: \ (1.7k)\,(16)} \\
- Mean Offset  &  86.2$_{\pm1.2}$ & 84.6$_{\pm1.3}$ & 78.0$_{\pm3.2}$ & \textbf{83.0} &  88.1$_{\pm1.2}$ & 85.6$_{\pm1.1}$ & 83.0$_{\pm3.3}$ & \textbf{85.6} \\
- Mean Offset + Regression  & 86.1$_{\pm1.2 }$& 84.1$_{\pm1.3}$ & 78.2$_{\pm3.1}$ & 82.8 & 88.3$_{\pm1.0}$ & 85.2$_{\pm1.5}$ & 83.4$_{\pm3.3}$ & \textbf{85.6} \\
\bottomrule
\end{tabular}
\caption{\textbf{Baseline trained with all original samples}: A comparison of our offset-based models to the \emph{Original} classifier trained on \emph{all} 24k original samples.
}\label{table:more_original}
\end{table*}

\begin{figure*}
    \centering
    \begin{subfigure}{0.25\textwidth}
      \includegraphics[height=3.5cm,width=\textwidth]{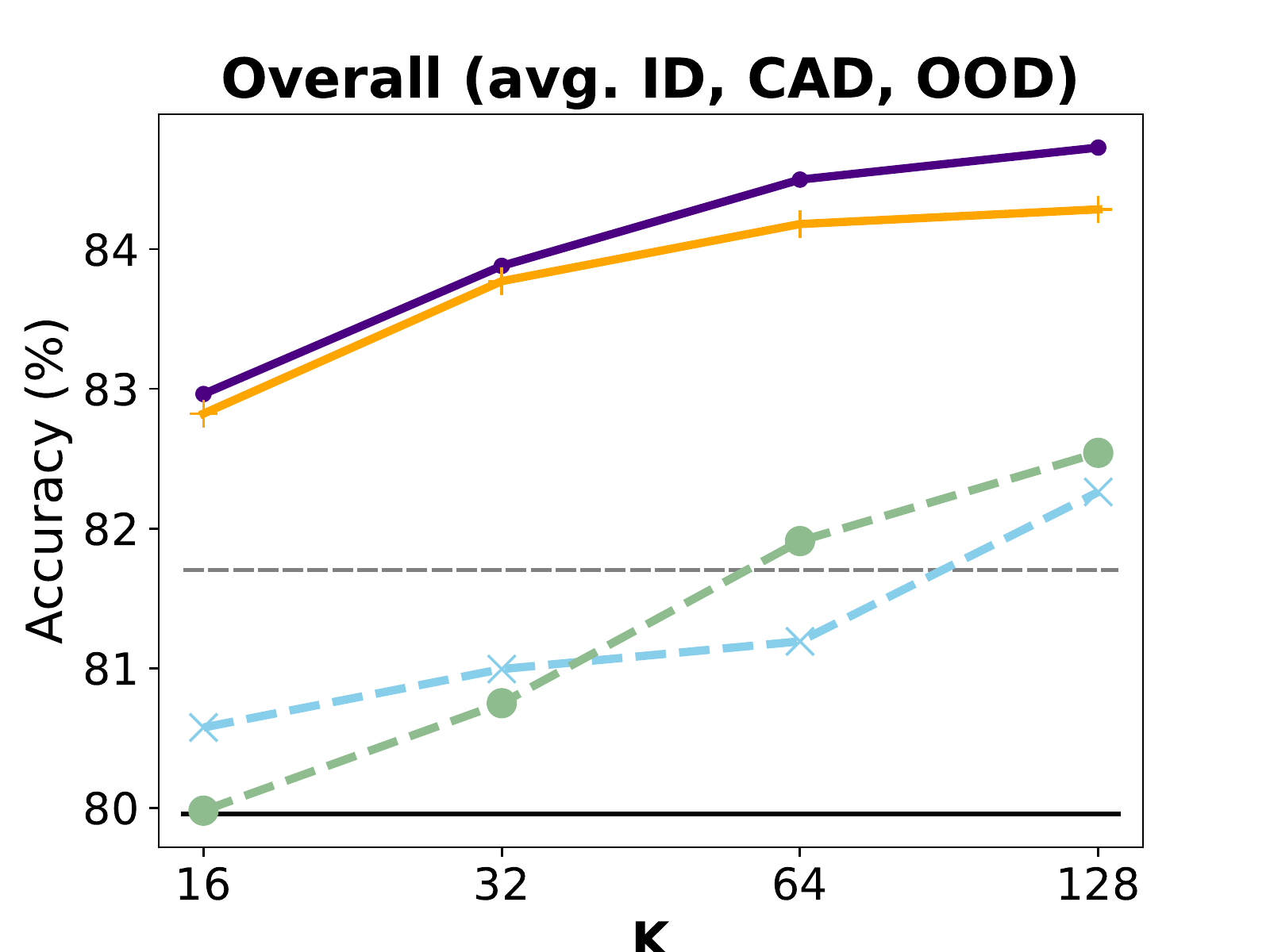}
    \end{subfigure}\hspace{-1em}
    \begin{subfigure}{0.25\textwidth}
      \includegraphics[height=3.5cm, width=\textwidth]{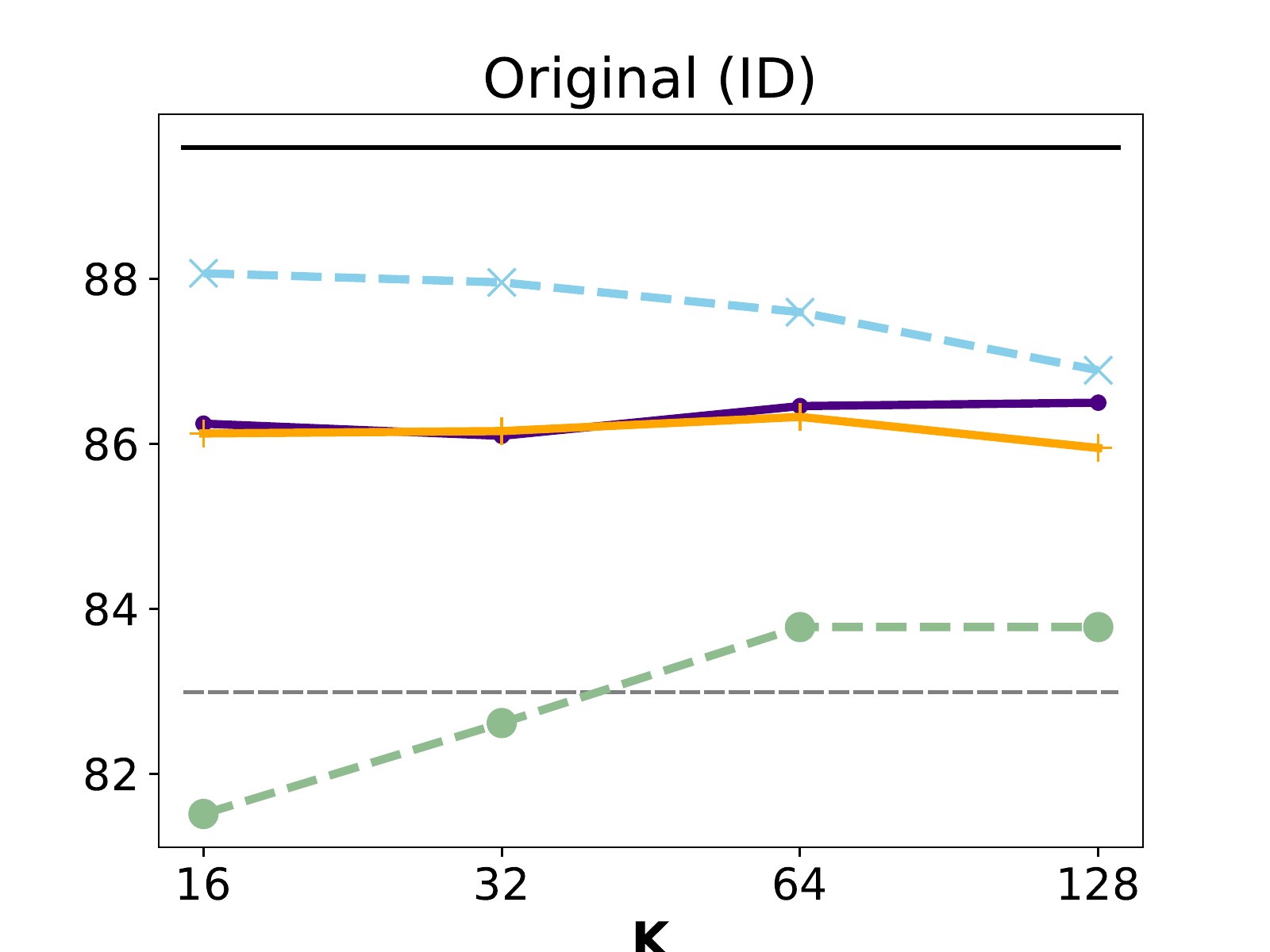}
    \end{subfigure}\hspace{-1em}
    \begin{subfigure}{0.25\textwidth}
      \includegraphics[height=3.5cm,width=\textwidth]{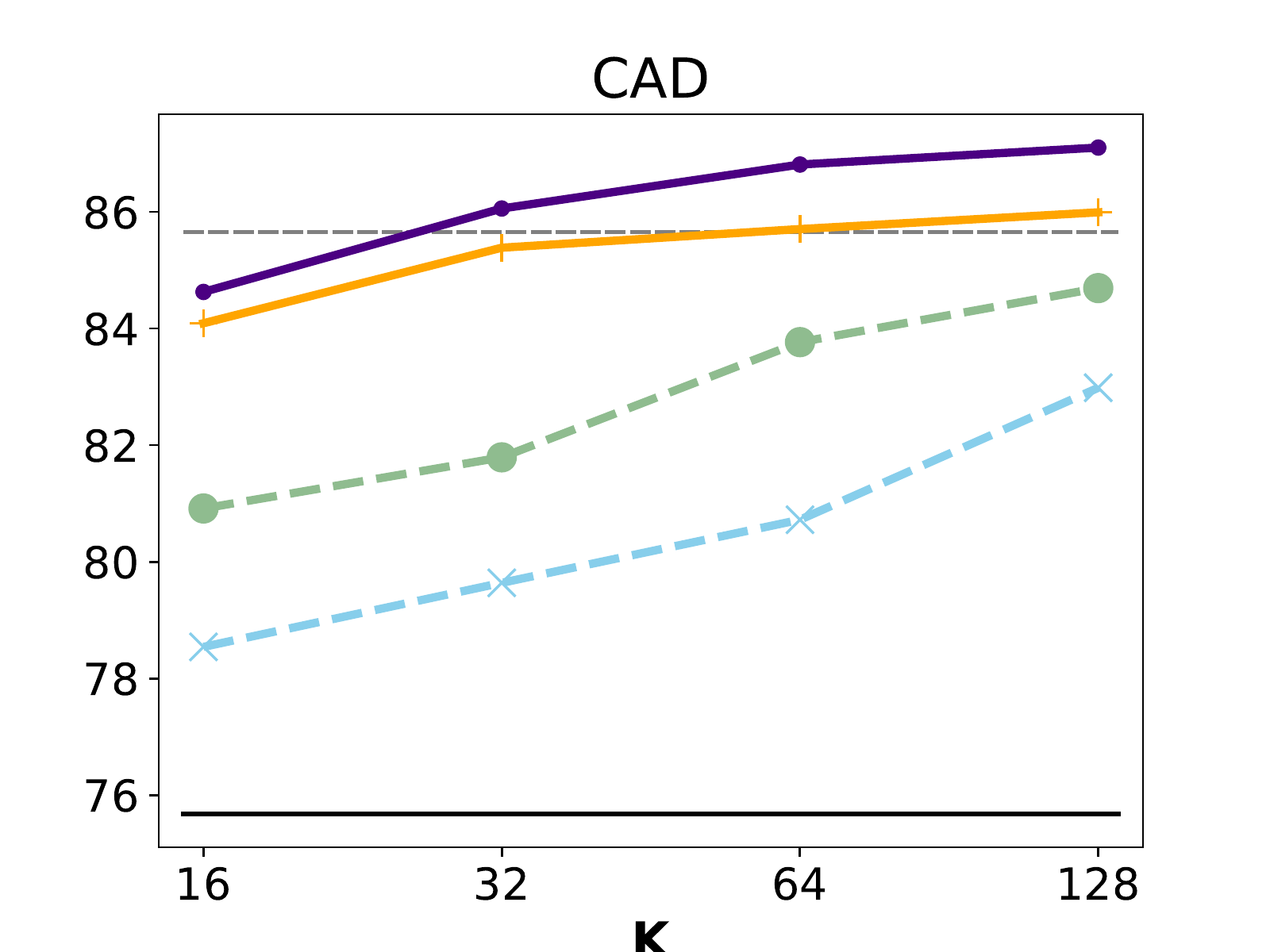}
    \end{subfigure}\hspace{-1em}
    \begin{subfigure}{0.25\textwidth}
      \includegraphics[height=3.5cm,width=\textwidth]{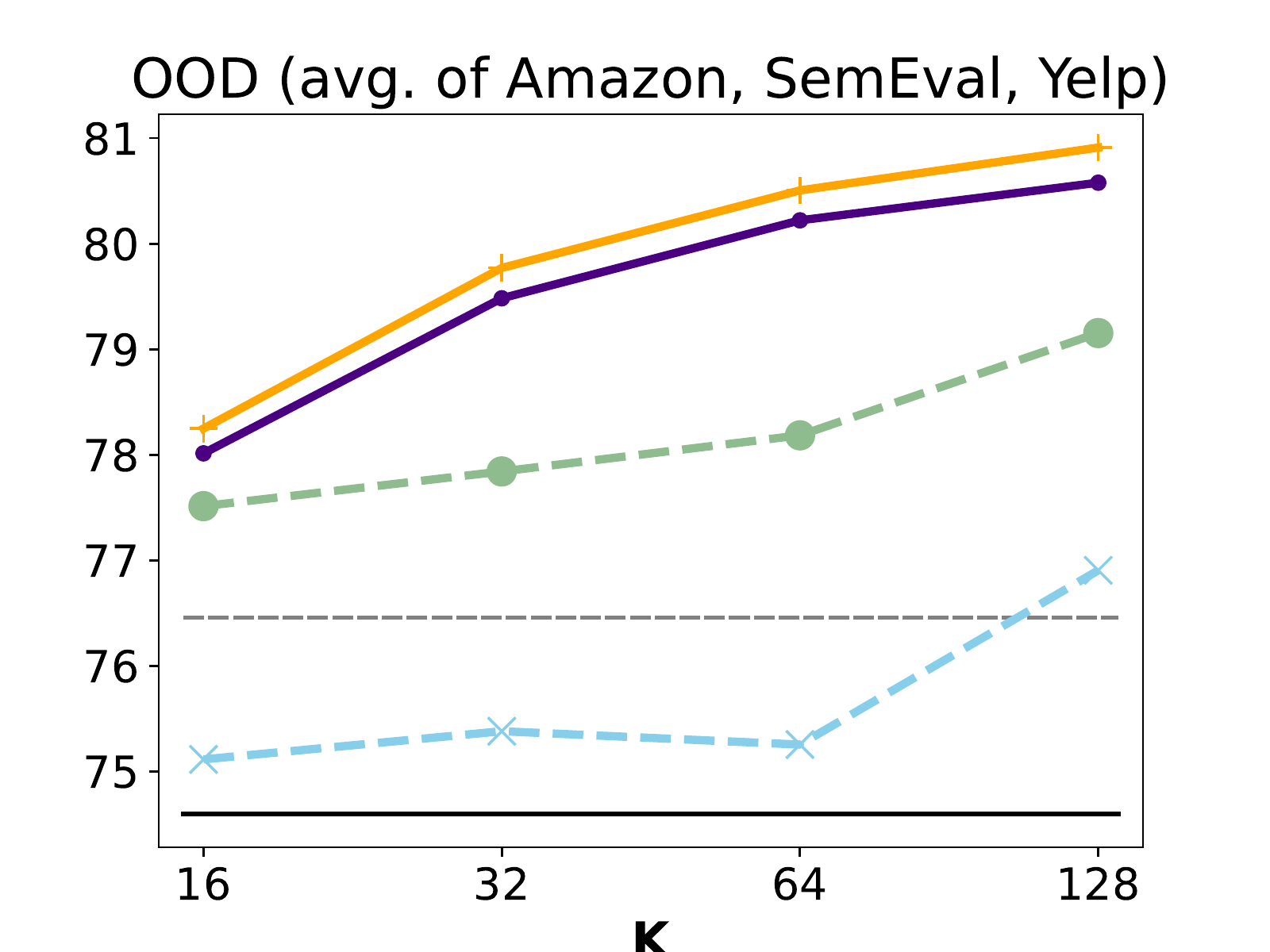}
    \end{subfigure}
    \begin{subfigure}{1.\textwidth}
      \includegraphics[width=\textwidth]{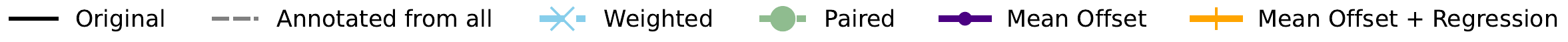}
       \label{fig:legend-simcse-roberta-large}
    \end{subfigure}\vspace{-1.5em}
    \caption{SimCSE-RoBERTa\textsubscript{\texttt{large}}} 
    \label{fig:sroberta-large}
\end{figure*}

\begin{figure*}
    \centering
    \begin{subfigure}{0.25\textwidth}
      \includegraphics[height=3.5cm,width=\textwidth]{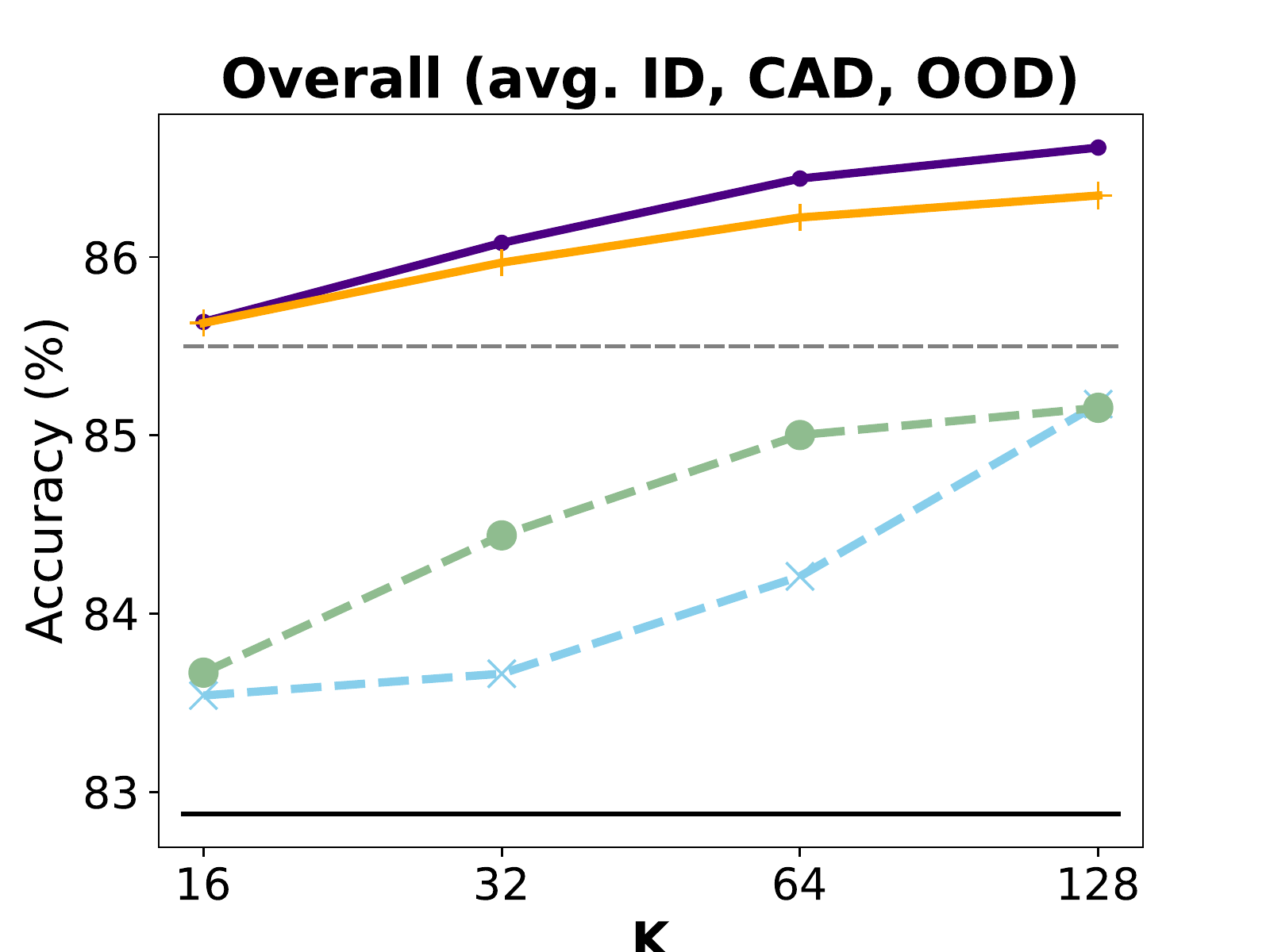}
    \end{subfigure}\hspace{-1em}
    \begin{subfigure}{0.25\textwidth}
      \includegraphics[height=3.5cm, width=\textwidth]{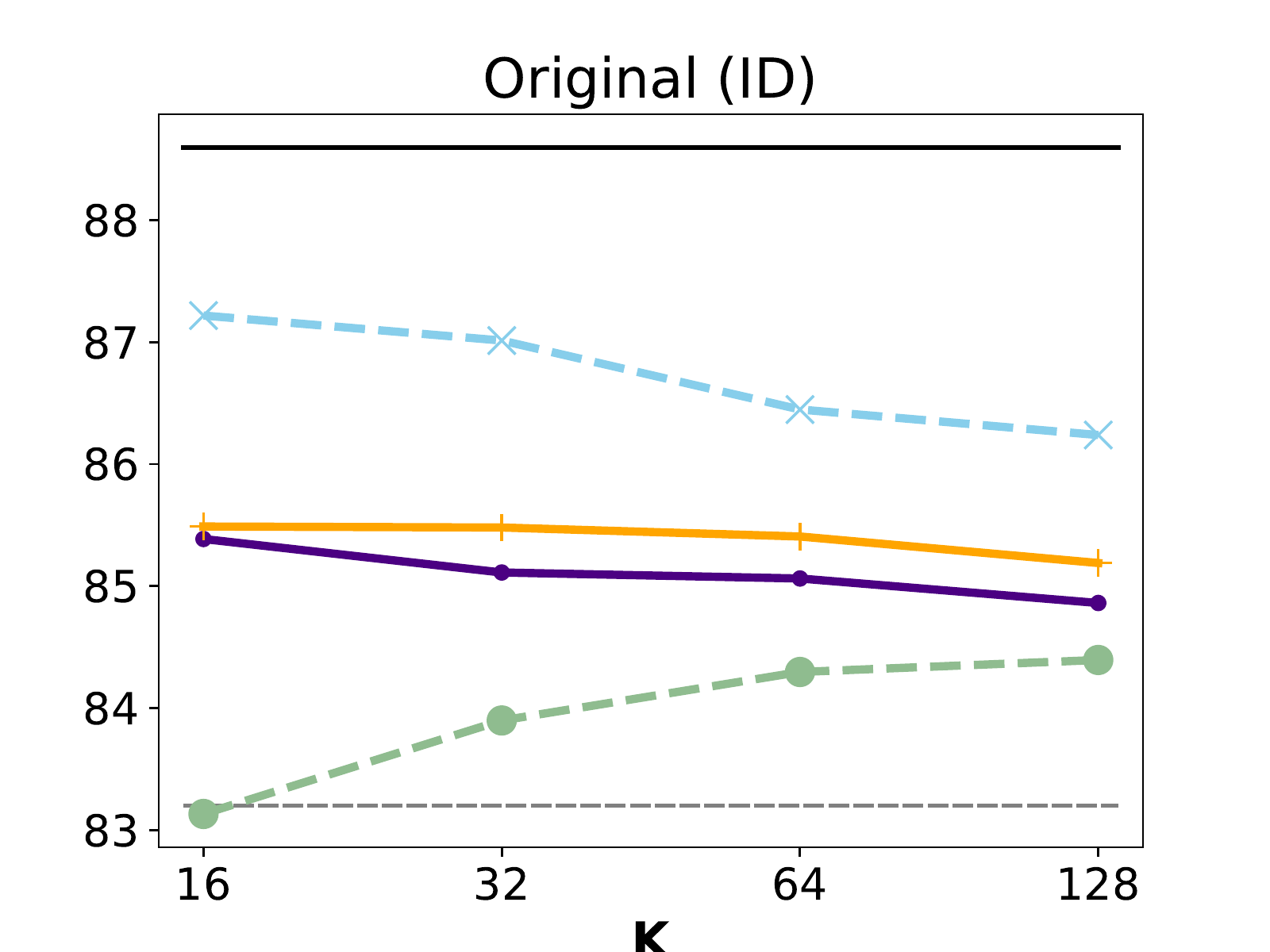}
    \end{subfigure}\hspace{-1em}
    \begin{subfigure}{0.25\textwidth}
      \includegraphics[height=3.5cm,width=\textwidth]{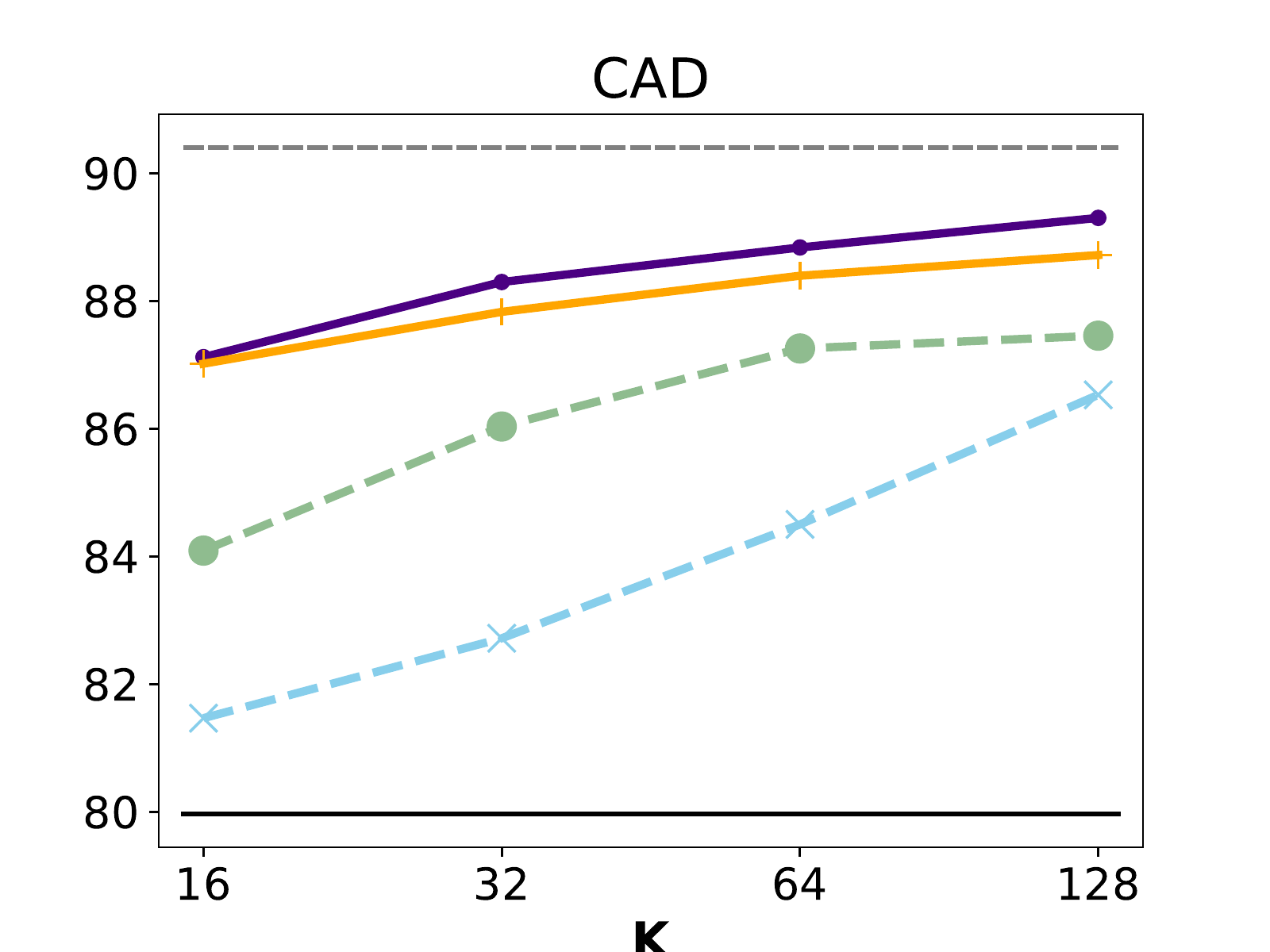}
    \end{subfigure}\hspace{-1em}
    \begin{subfigure}{0.25\textwidth}
      \includegraphics[height=3.5cm,width=\textwidth]{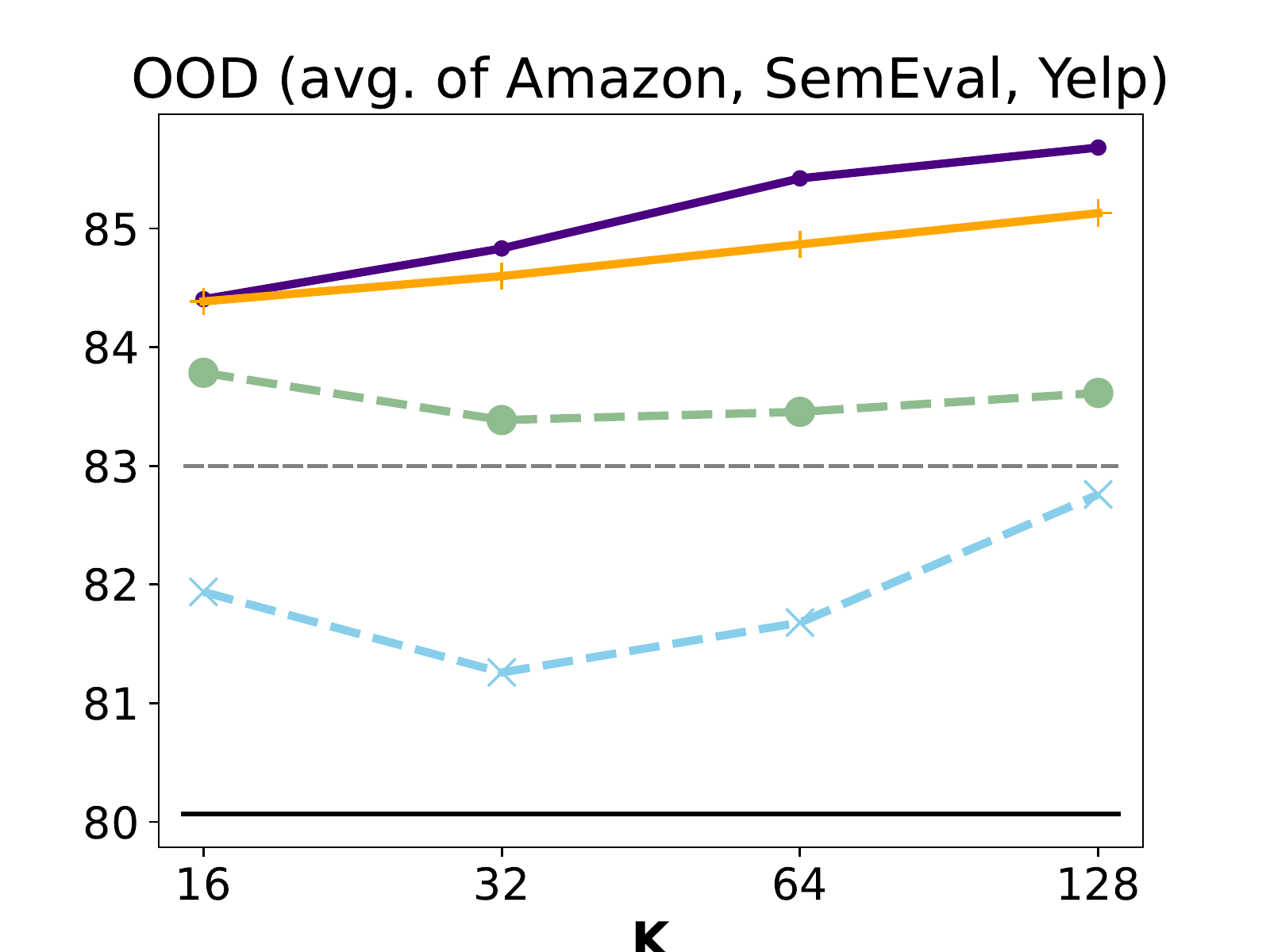}
    \end{subfigure}
    \begin{subfigure}{1.\textwidth}
      \includegraphics[width=\textwidth]{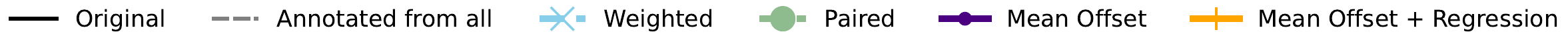}
       \label{fig:legend-unsup-simcse-bert-large}
    \end{subfigure}\vspace{-1.5em}
    \caption{SimCSE-BERT\textsubscript{\texttt{large}}} 
    \label{fig:unsup-simcse-bert-large}
\end{figure*}

\begin{figure*}
    \centering
    \begin{subfigure}{0.25\textwidth}
      \includegraphics[height=3.5cm,width=\textwidth]{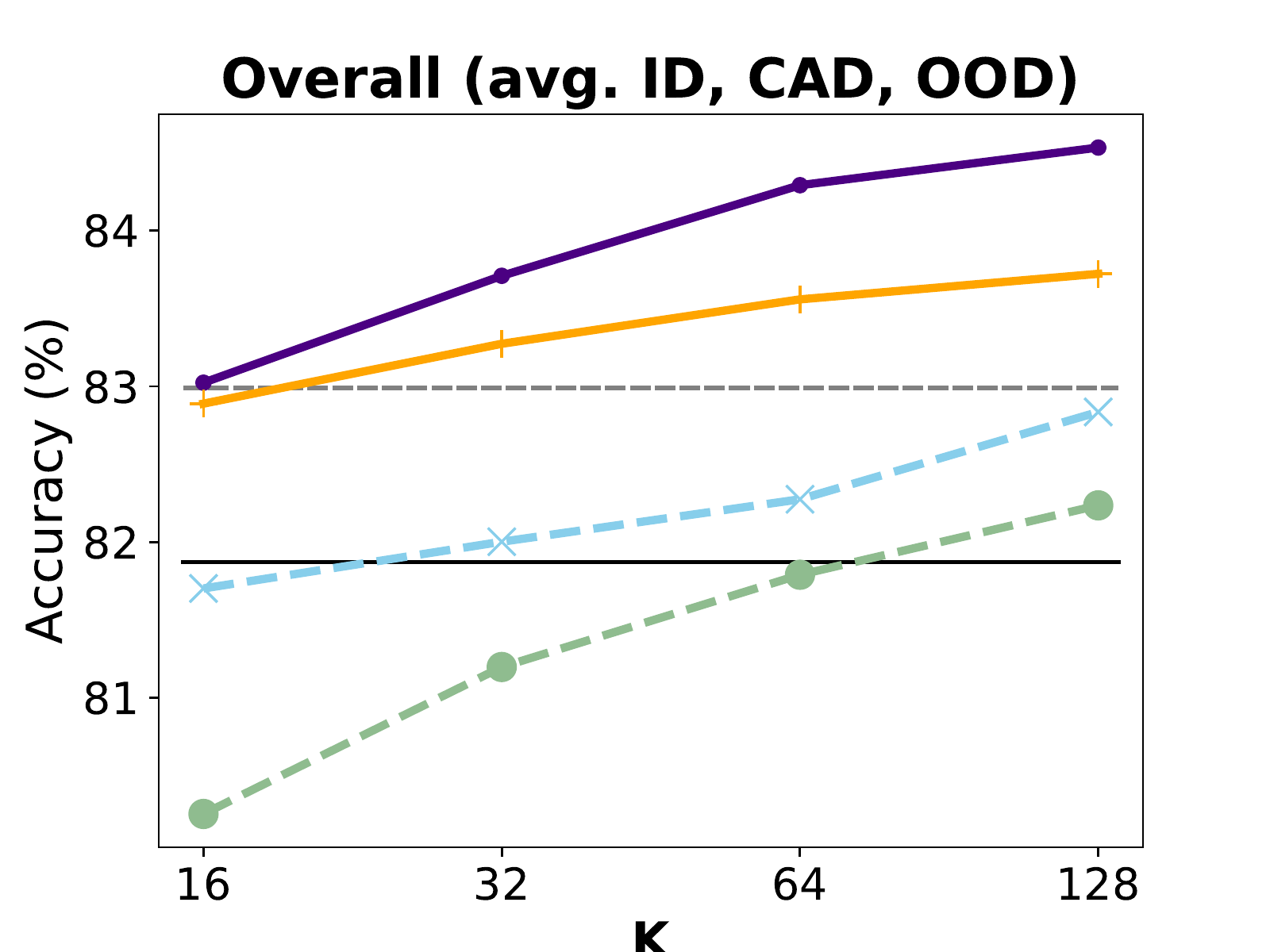}
    \end{subfigure}\hspace{-1em}
    \begin{subfigure}{0.25\textwidth}
      \includegraphics[height=3.5cm, width=\textwidth]{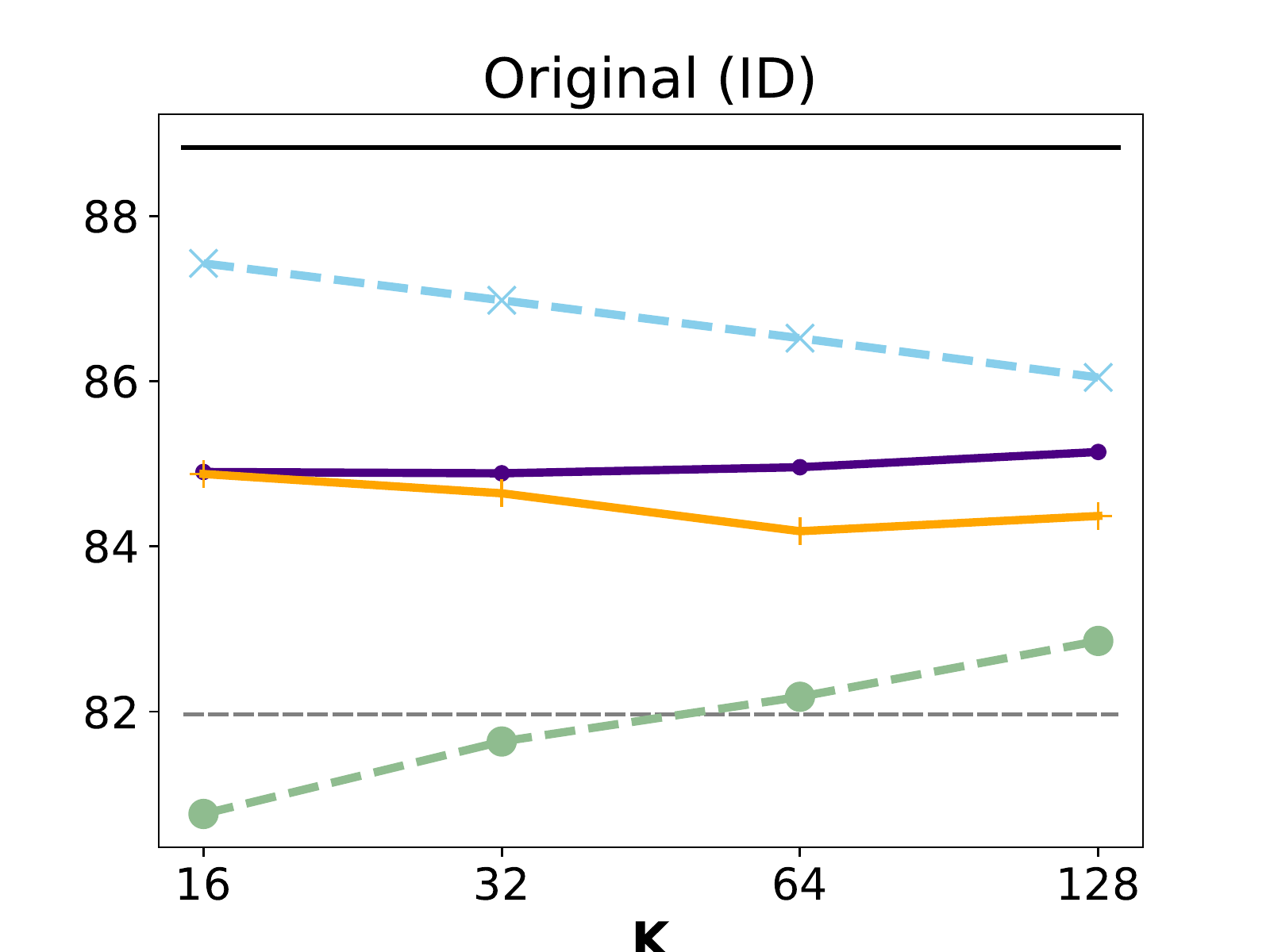}
    \end{subfigure}\hspace{-1em}
    \begin{subfigure}{0.25\textwidth}
      \includegraphics[height=3.5cm,width=\textwidth]{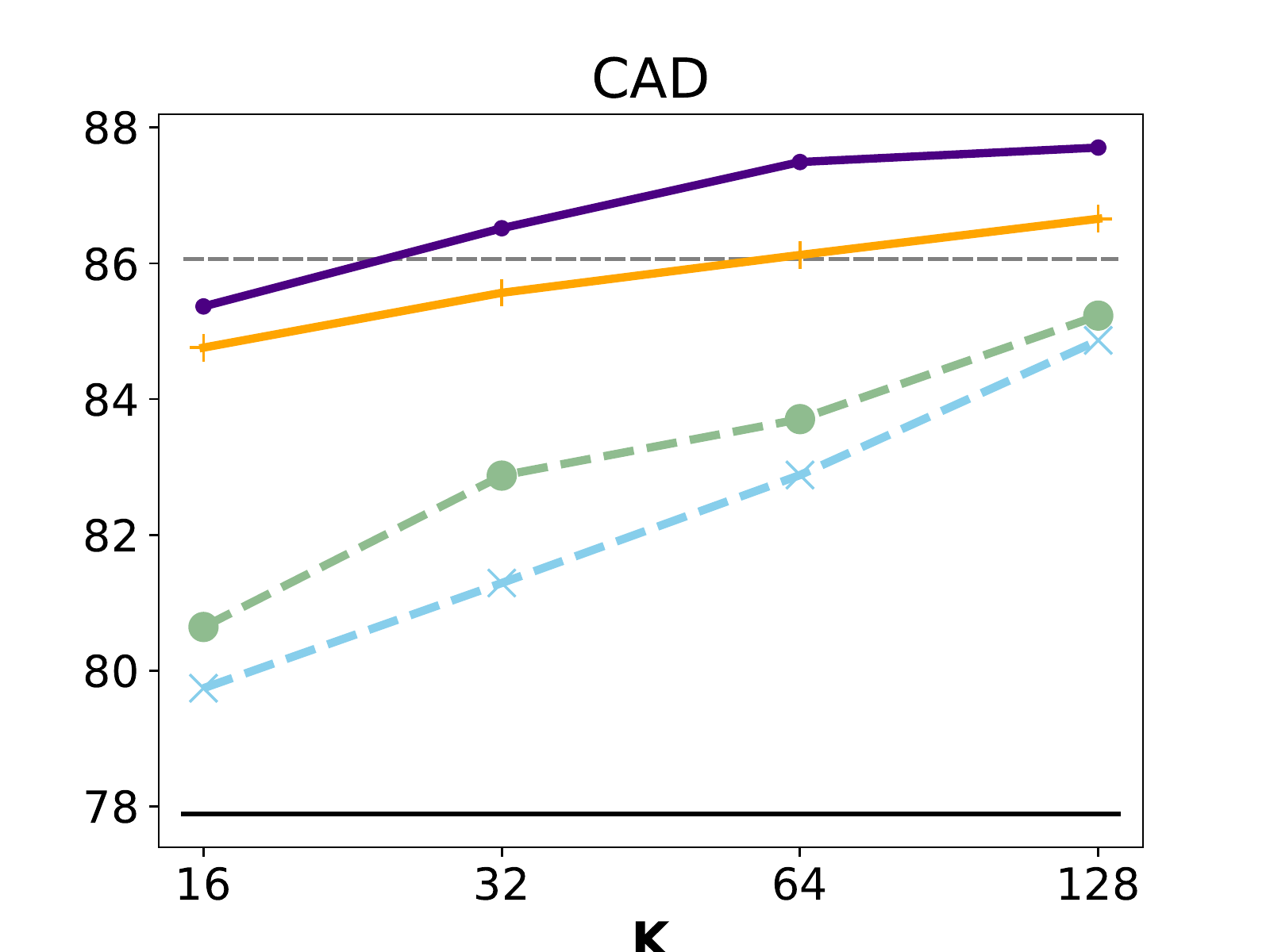}
    \end{subfigure}\hspace{-1em}
    \begin{subfigure}{0.25\textwidth}
      \includegraphics[height=3.5cm,width=\textwidth]{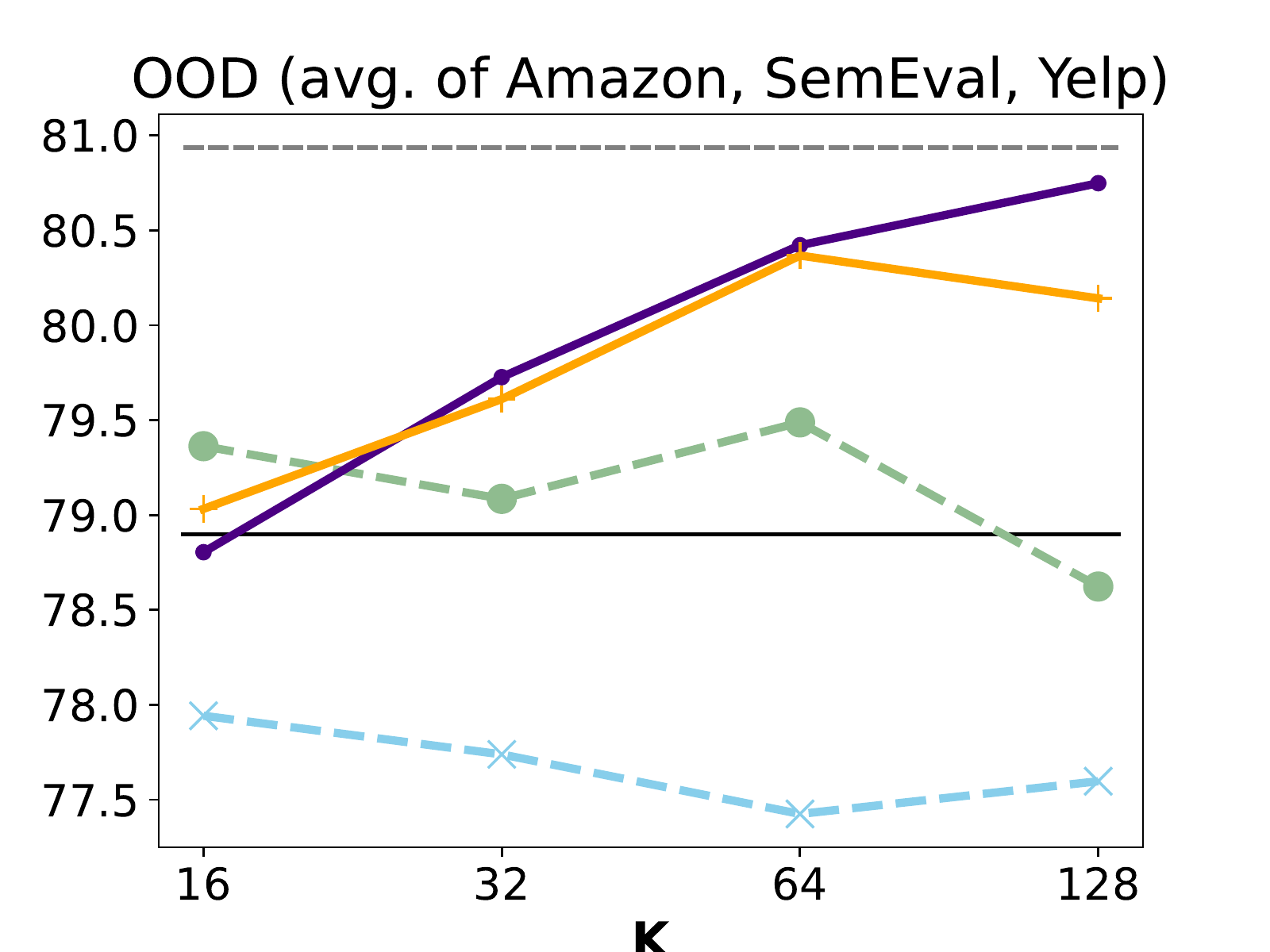}
    \end{subfigure}
    \begin{subfigure}{1.\textwidth}
      \includegraphics[width=\textwidth]{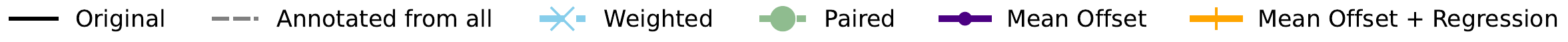}
       \label{fig:legend-bert-base}
    \end{subfigure}\vspace{-1.5em}
    \caption{SimCSE-BERT\textsubscript{\texttt{base}} } 
    \label{fig:unsup-simcse-bert-base}
\end{figure*}

\begin{figure*}
    \centering
    \begin{subfigure}{0.25\textwidth}
      \includegraphics[height=3.5cm,width=\textwidth]{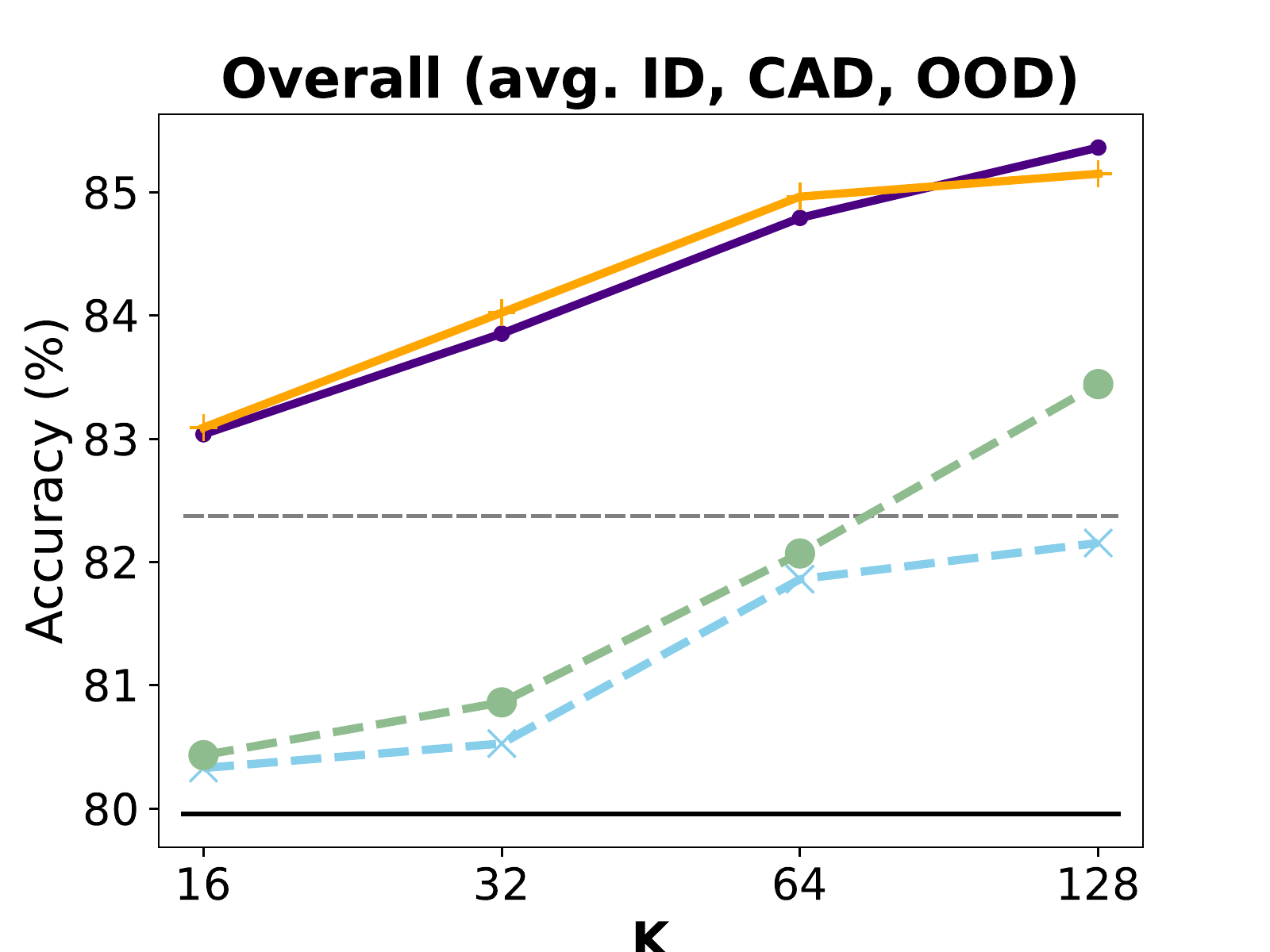}
    \end{subfigure}\hspace{-1em}
    \begin{subfigure}{0.25\textwidth}
      \includegraphics[height=3.5cm, width=\textwidth]{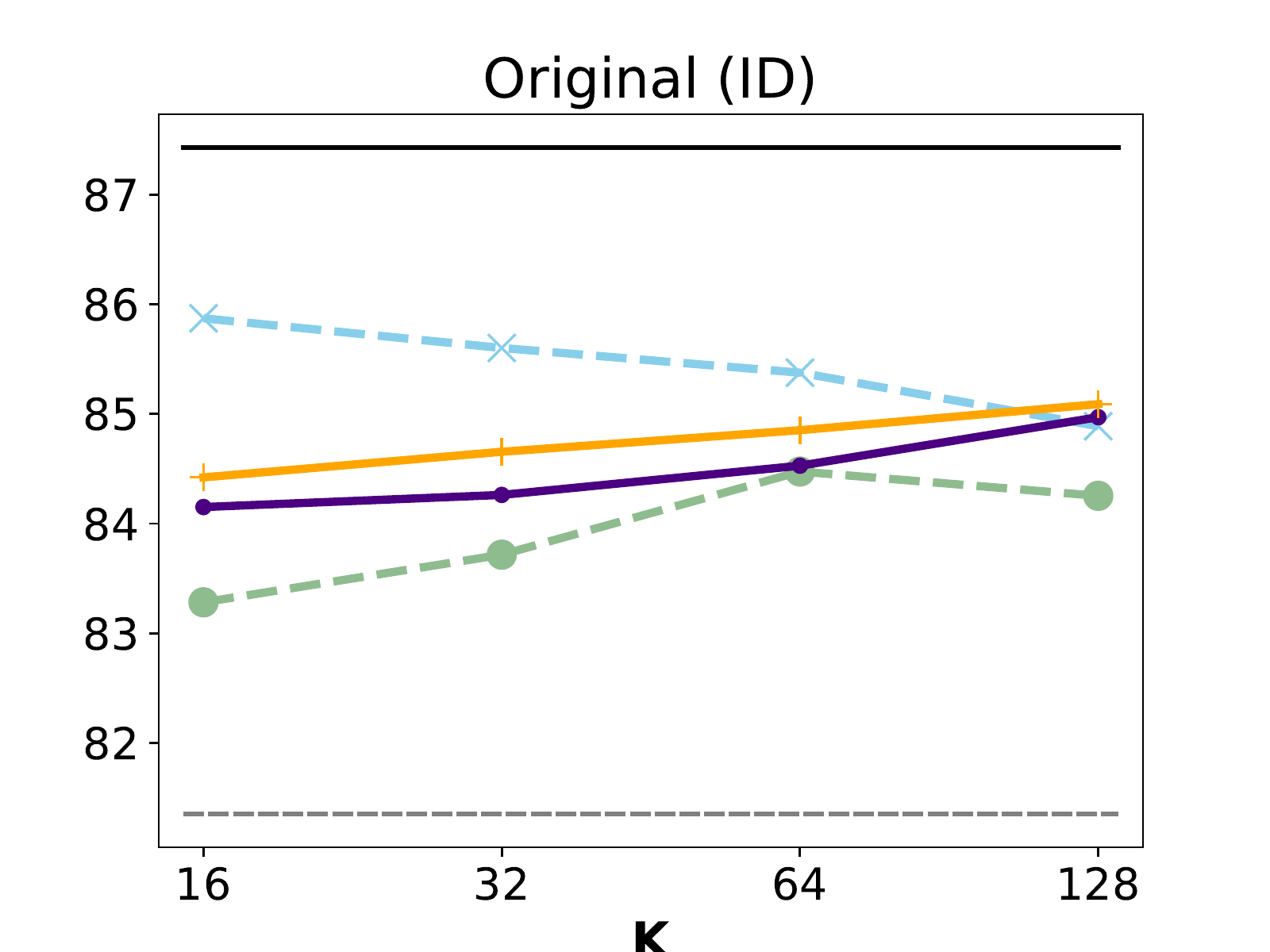}
    \end{subfigure}\hspace{-1em}
    \begin{subfigure}{0.25\textwidth}
      \includegraphics[height=3.5cm,width=\textwidth]{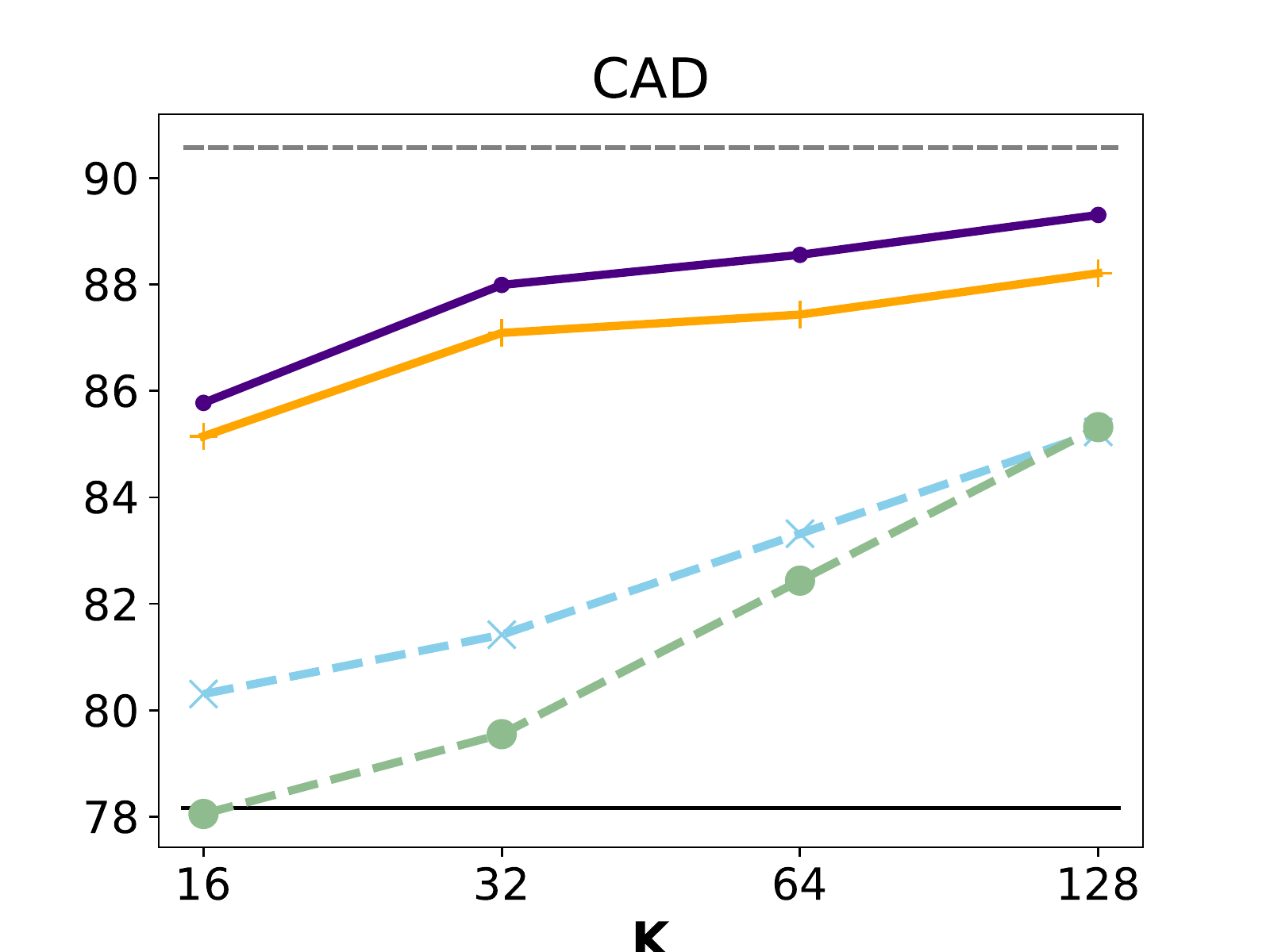}
    \end{subfigure}\hspace{-1em}
    \begin{subfigure}{0.25\textwidth}
      \includegraphics[height=3.5cm,width=\textwidth]{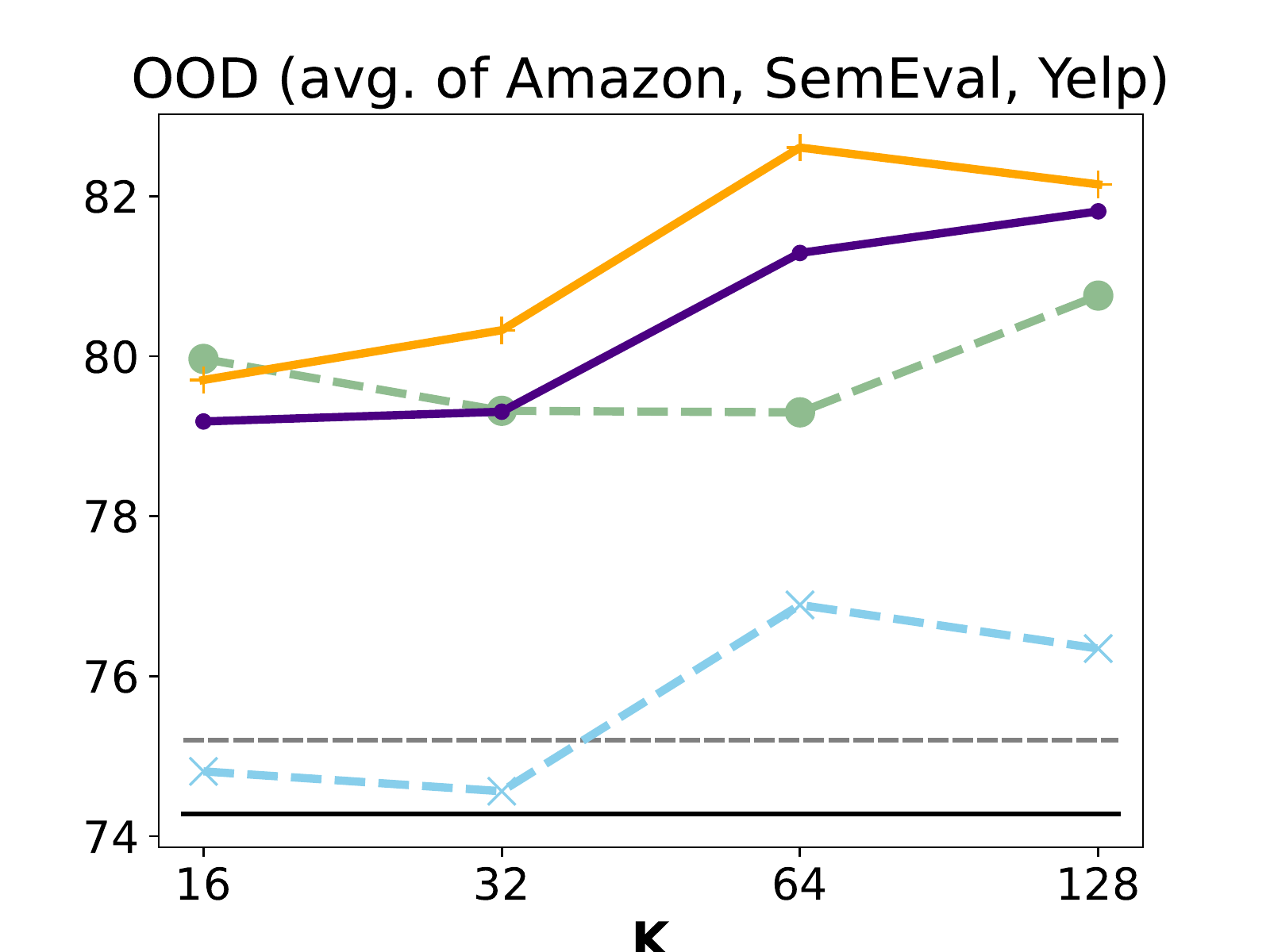}
    \end{subfigure}
    \begin{subfigure}{1.\textwidth}
      \includegraphics[width=\textwidth]{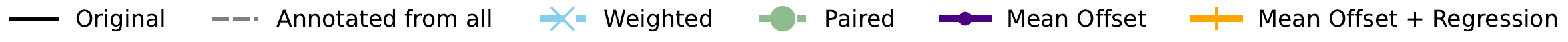}
       \label{fig:legend-sdistillroberta}
    \end{subfigure}\vspace{-1.5em}
    \caption{SDistillRoBERTa} 
    \label{fig:all-distilroberta-v1}
\end{figure*}

\begin{figure*}
    \centering
    \begin{subfigure}{0.25\textwidth}
      \includegraphics[height=3.5cm,width=\textwidth]{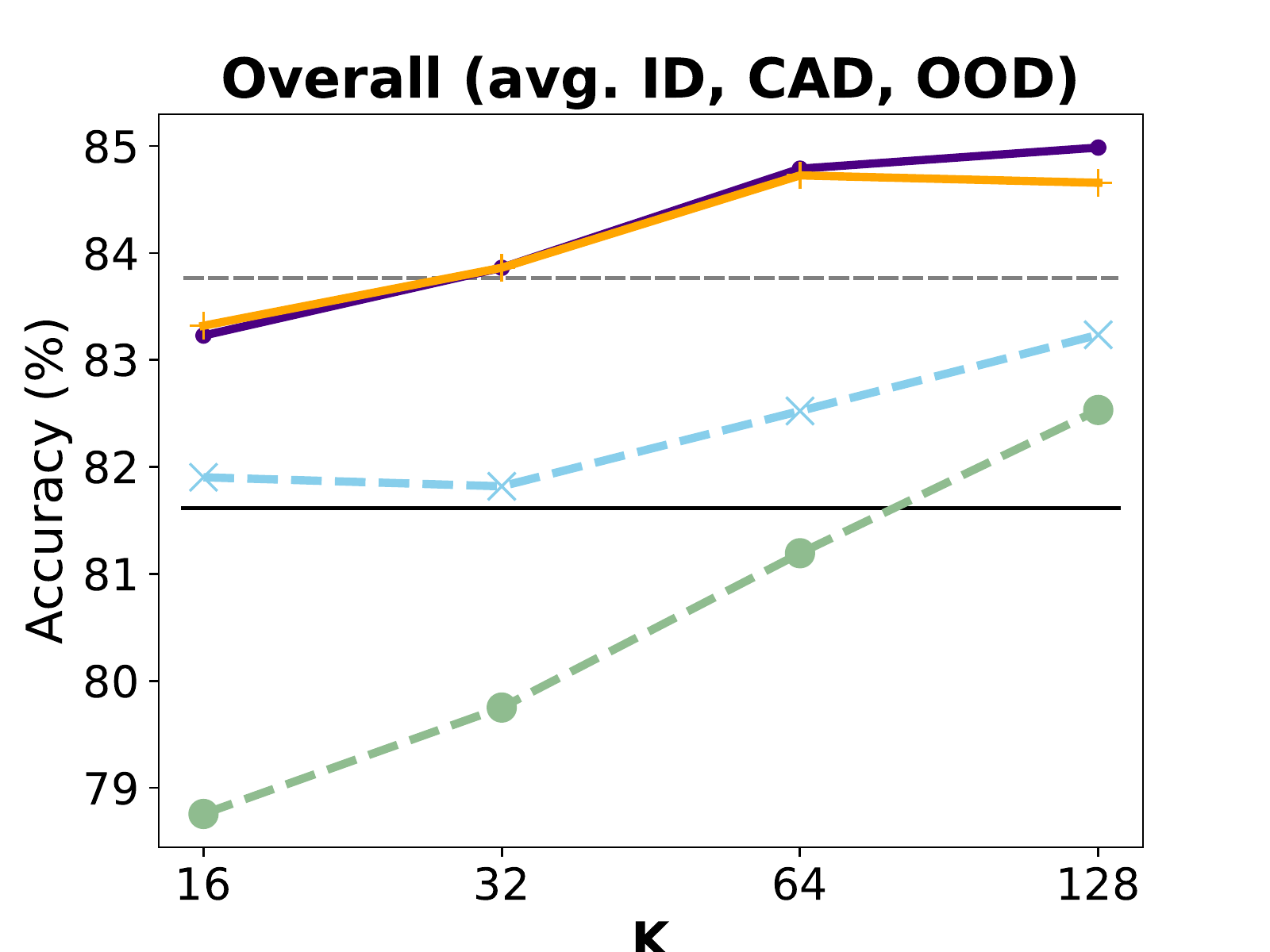}
    \end{subfigure}\hspace{-1em}
    \begin{subfigure}{0.25\textwidth}
      \includegraphics[height=3.5cm, width=\textwidth]{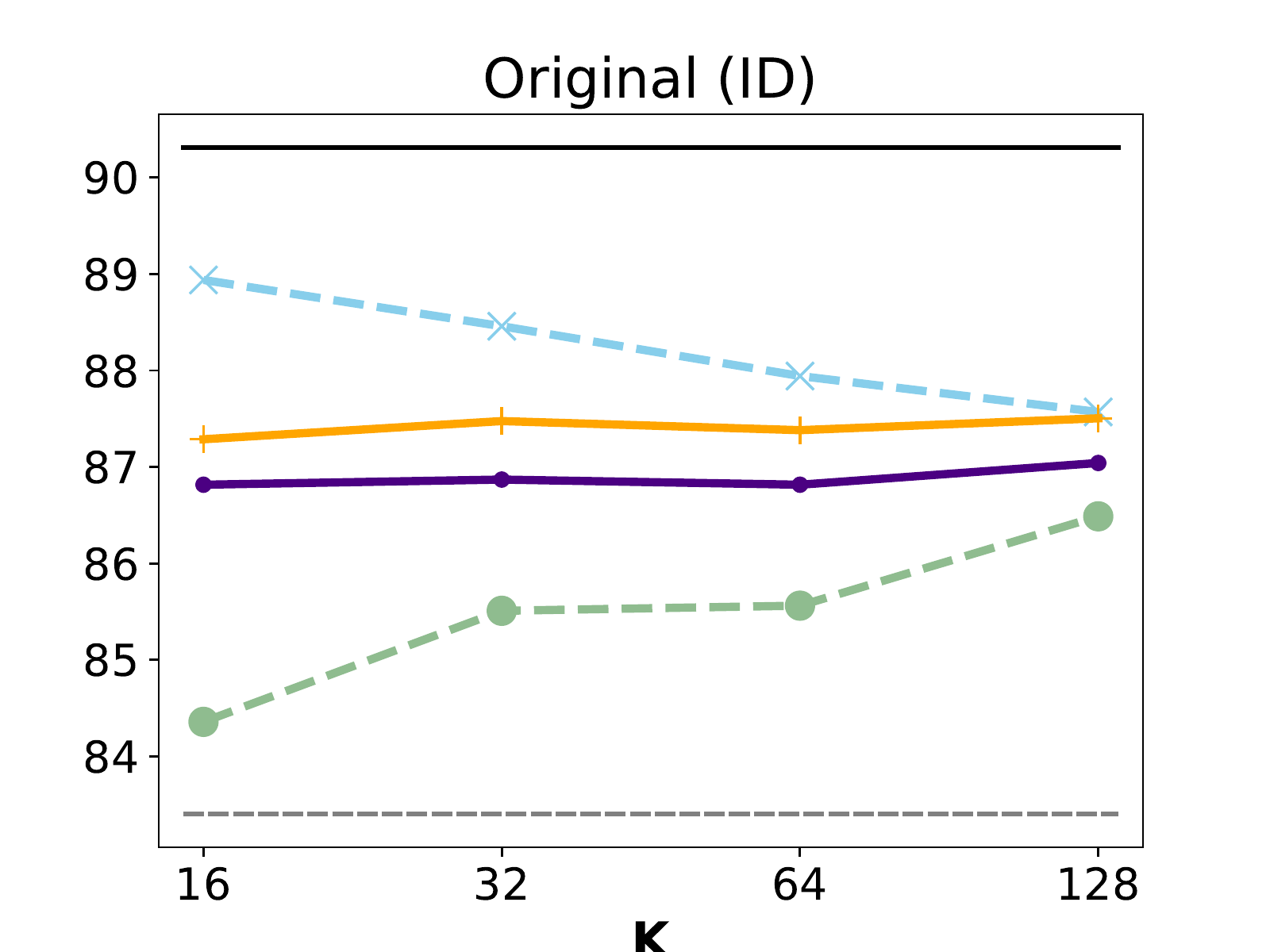}
    \end{subfigure}\hspace{-1em}
    \begin{subfigure}{0.25\textwidth}
      \includegraphics[height=3.5cm,width=\textwidth]{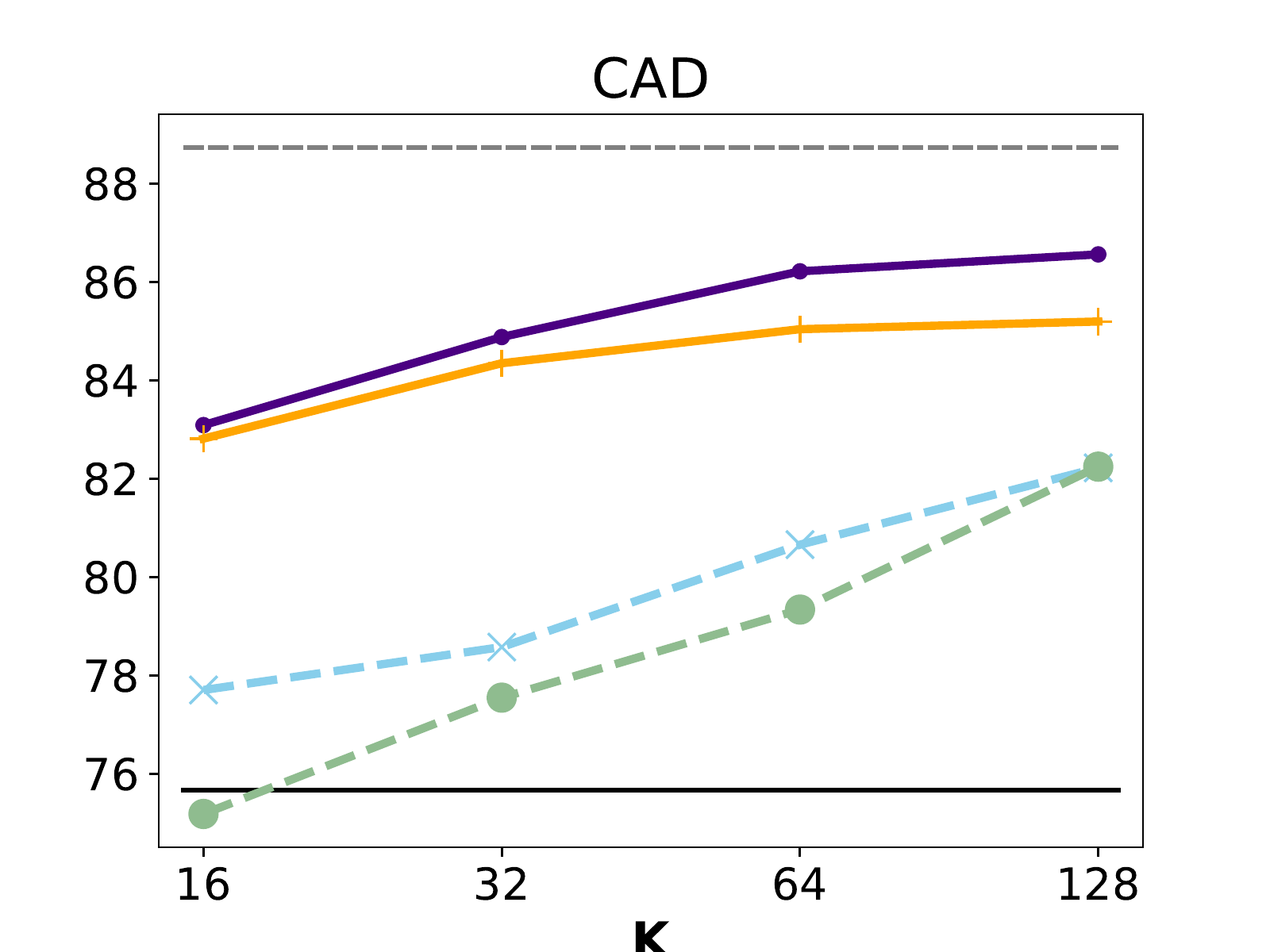}
    \end{subfigure}\hspace{-1em}
    \begin{subfigure}{0.25\textwidth}
      \includegraphics[height=3.5cm,width=\textwidth]{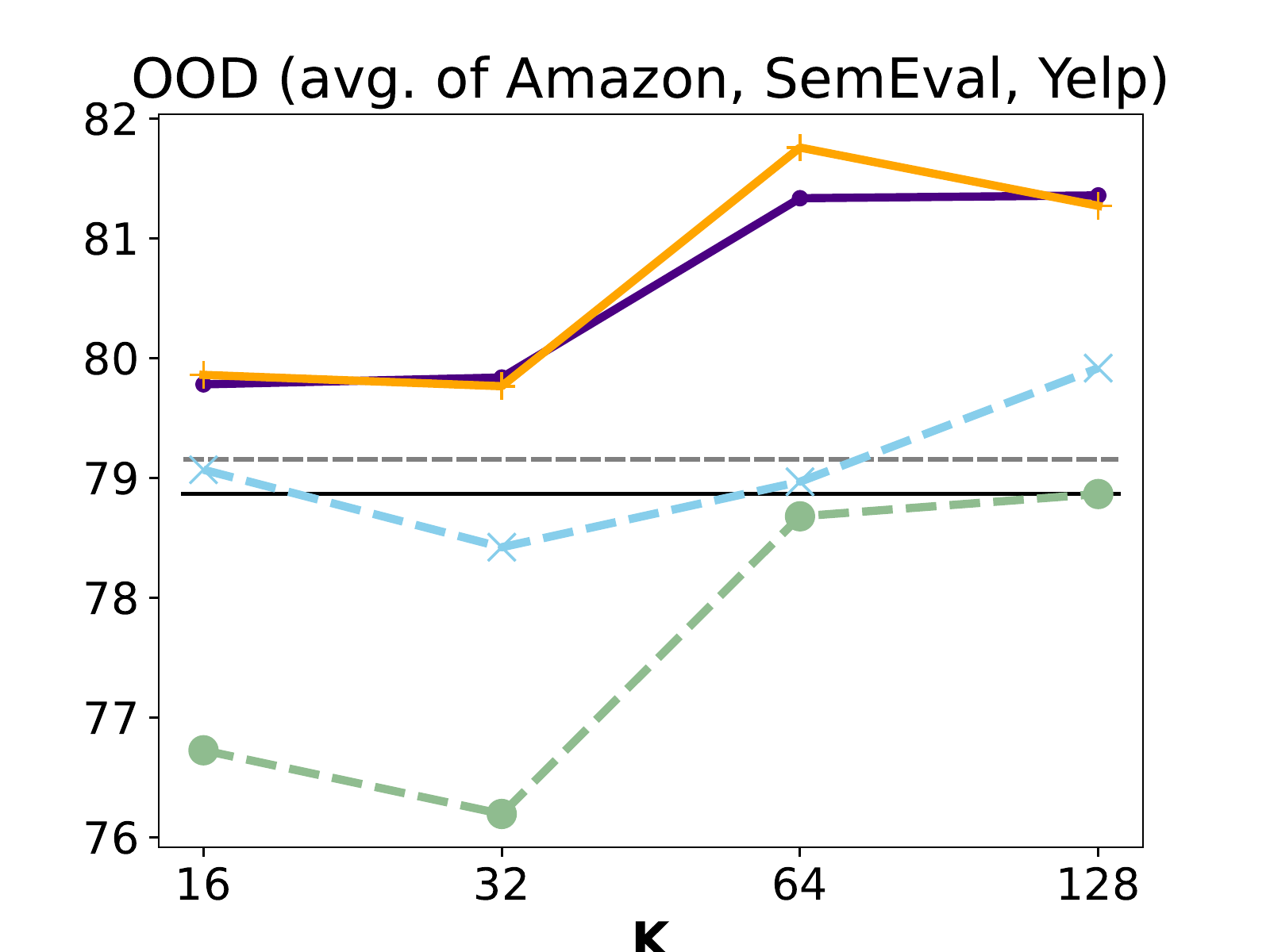}
    \end{subfigure}
    \begin{subfigure}{1.\textwidth}
      \includegraphics[width=\textwidth]{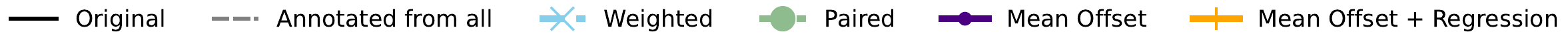}
       \label{fig:legend-sdistillroberta}
    \end{subfigure}\vspace{-1.5em}
    \caption{SMPNet} 
    \label{fig:smpnet}
\end{figure*}

From the leftmost graphs in Figs.~\ref{fig:sroberta-large}--\ref{fig:smpnet}, we observe that the different classifiers follow the main trends as discussed in \secref{sec:results}: the models trained on generated counterfactual vectors from the mean offset models, are overall most robust and outperform both \begin{enumerate*}[(i)]
    \item classifiers without generated counterfactuals (\emph{Weighted} and \emph{Paired}) and
    \item classifiers trained on counterfactuals generated from the best model of \citet{wang2021robustness}.
\end{enumerate*}
This holds for all sentence encoders and values of $k \in \lbrace$16, 32, 64, 128$\rbrace$, with the sole exception of SMPNet and SimCSE-BERT$_{\texttt{base}}$ for which \emph{Annotated from all} (based on \citet{wang2021robustness}) is slightly better than the offset-based models for $k=$~16, but still worse when $k>$~16.
Hence, we stand by the main paper's stated conclusions.

Furthermore, Table 5 highlights that classifiers trained on our offset-based counterfactuals (using 1.7k original and just $k=$~16 manual counterfactuals) can outperform the \emph{Original} baseline trained on \emph{all} 24k original samples  (most clear for SRoBERTa\textsubscript{\texttt{large}}). It is worth noting that the \emph{Original} baseline becomes more robust when trained on all 24k original samples rather than 3.4k samples. However, to make the \emph{Original} classifier more robust requires annotating 22.3k extra in-distribution samples, rather than counterfactually revising only 16 original samples.

\subsection{Strengths and Weaknesses}\label{subsec:individual_robustness} Below we discuss the strengths and weaknesses of the different approaches by considering all sentence encoders.

\paragraph{Paired} Following the results in \secref{sec:results}, the \emph{Paired} approach consistently yields high accuracies on the out-of-distribution (OOD) test set: for all encoders we observe a notable improvement over the \emph{Original} classifier, with SMPNet as the exception.
Moreover, the \emph{Paired} model reaches similar or even slightly better OOD performance compared to the best approaches for small values $k \in \lbrace$16, 32$\rbrace$.
However, when evaluated on in-distribution (ID) samples, the \emph{Paired} model degrades significantly in accuracy compared to the \emph{Original} classifier and the majority of all other approaches.\footnote{With the exception of particularly \citet{wang2021robustness}, for larger $k$ values and all encoders.}
For counterfactuals (CAD), the \emph{Paired} model improves upon the \emph{Original} classifier but performs worse than the approaches that train with generated counterfactuals (\ie ours and \citet{wang2021robustness}).

\paragraph{Weighted} We observe similar trends as discussed in \secref{sec:results} where the \emph{Weighted} model retains most of the \emph{Original} classifier's performance on  in-distribution samples, and more so than any of the other approaches. While \emph{Weighted} performs better than the \emph{Original} model on CAD it  performs significantly worse than the classifiers trained with generated counterfactuals. The generalization of \emph{Weighted} to out-of-distribution is mixed, where only for some encoders it yields slightly better results than the \emph{Original} classifier, but it is consistently worse when compared to the \emph{Paired} model and classifiers trained with generated counterfactuals from our models.

\paragraph{Annotated from top} \cite{wang2021robustness}  The results for other sentence encoders, again, are similar to those reported in \secref{sec:results}. The classifier trained with counterfactuals from this model degrades significantly on in-distribution performance, more than any of the other approaches except for SimCSE-BERT\textsubscript{\texttt{base}} and SimCSE-BERT\textsubscript{\texttt{large}}, where the \emph{Paired} model performs worse for $k<$~64.
For CAD, this model is the most accurate compared to the other approaches (with $k=$~16).
Except for SimCSE-BERT\textsubscript{\texttt{base}}, the classifiers tends to generalize worse for OOD-samples compared to our approaches that train on generated counterfactuals, and for SRoBERTa\textsubscript{\texttt{large}} worse than the \emph{Original} classifier. 

\paragraph{Mean Offset (+Regression)} The classifiers trained with counterfactuals generated from our models slightly drop in accuracy on in-distribution samples but perform better, except the \emph{Weighted} model, than all the other approaches.
On CAD, the classifiers perform either best or come in second to the best model from \cite{wang2021robustness}.
Similarly, they perform best or come in second to the \emph{Paired} model for OOD (except for SimCSE-BERT\textsubscript{\texttt{base}} for which \cite{wang2021robustness} is better). Note that our offset-based models, with a slight drop in in-distribution accuracy but consistent improvements on both CAD and OOD data, results in classifiers that strike the desirable balance across the three different test distributions (ID, CAD, OOD).

\section{Experimental Details}\label{app:implementation_details}

\begin{table}
\small
\centering
\addtolength{\tabcolsep}{-1pt}
\begin{tabular}{lcc}
\toprule
Dataset & \#\,Documents & \#\,Tokens (avg.) \\
\midrule
\multicolumn{3}{l}{\textbf{In-distribution}} \\
- IMDb train ($\Dset{train}{ID}$)  & 1,707 & 163 \\
- IMDb test ($\Dset{test}{ID}$)  & 488 & 162 \\
\midrule
\multicolumn{3}{l}{\textbf{Counterfactual}} \\
- IMDb train ($\Dset{train}{CAD}$)  & 1,707 & 162 \\
- IMDb test ($\Dset{test}{CAD}$)  & 488 & 162 \\
\midrule
\multicolumn{3}{l}{\textbf{Out-of-distribution}} \\
 - Amazon test ($\Damzn$) & 5,766 & 132 \\
 - Yelp test ($\Dyelp$) & 6,462 & 120 \\
 - SemEval test ($\Dsemeval$) & 130,126 & 20 \\
\bottomrule
\end{tabular}
\addtolength{\tabcolsep}{1pt}
\caption{Dataset statistics}
\label{table:datasets}
\end{table}

\paragraph{Datasets} \tabref{table:datasets} summarizes the dataset statitics, reporting per dataset/split:
\begin{enumerate*}[(i)]
\item the number of documents and 
\item the average number of tokens per document.
\end{enumerate*}
All datasets are equally balanced between the positive and negative classes.

\paragraph{Sentence Encoders} 
For the SBERT\footnote{\url{https://www.sbert.net}} architecture, we reported results for models based on RoBERTa \cite{liu2019roberta}, DistilRoBERTa, and MPNet \cite{NEURIPS2020_c3a690be} with corresponding Hugging Face \cite{wolf-etal-2020-transformers} identifiers: \emph{all-roberta-large-v1}, \emph{all-distilroberta-v1} and \emph{all-mpnet-base-v2}.
\noindent For SimCSE,\footnote{\url{https://github.com/princeton-nlp/SimCSE}}  we experimented with models based on RoBERTa\textsubscript{\texttt{large}} \cite{liu2019roberta}, BERT\textsubscript{\texttt{large}} and based on BERT\textsubscript{\texttt{large}} \cite{devlin-etal-2019-bert}, with as  Hugging Face names \emph{unsup-simcse-roberta-large}, \emph{unsup-simcse-bert-large} and \emph{unsup-simcse-bert-base}.
Before training the linear classifiers on CPU, we pre-computed all the encodings for the different datasets on a single GeForce GTX 1080 Ti, taking at most one hour for each encoder. All sentence encoders yield vectors of dimension $d$ ranging 
between 768 and 1,024.

\paragraph{Linear Classifiers}
As stated before, classifiers in the experiments are trained with logistic regression and 4-fold cross-validation (to determine L2 regularization parameter $\lambda$) for which we used the \texttt{LogisticRegressionCV} implementation of Scikit-Learn \cite{pedregosa2011scikit}.
We choose the `lbfgs' solver, set the maximum number of iterations to 4,000, and used for the `Cs' parameter the inverse regularization values of those reported in the paper.
The classifiers can easily be trained and evaluated on all datasets on a 2,6 GHz 6-Core Intel Core i7, taking less than one minute per run. 

\paragraph{Linear Regression}
We implemented the \emph{Mean Offset + Regression} model using the \texttt{LinearRegression} implementation of Sklearn  with default parameters and ordinary least squares. Computing both the mean offset and the transformation gives negligible overhead and can be done within a fraction of a second on a on a 2,6~GHz 6-Core Intel Core i7 CPU.

\paragraph{Code} Our code and data to reproduce the experimental results is publicly available\footnote{\url{https://github.com/maarten-deraedt/EMNLP2022-robustifying-sentiment-classification}}.

\begin{table*}
\small
\centering
\begin{tabular}{lcccccccc}
\toprule
\multicolumn{9}{c}{SRoBERTa\textsubscript{\texttt{large}}}\\ 
& \multicolumn{4}{c}{\textbf{Free Reg.:} $\lambda \in \lbrace 10^{-3},10^{-2},\ldots, 10^3\rbrace$} & \multicolumn{4}{c}{\textbf{Strong Reg.:} $\lambda \in \lbrace 1, 10,\ldots, 10^3\rbrace$} \\ 
\cmidrule(lr){2-5} \cmidrule(lr){6-9}
 \textit{Model} $\ $ ($n$)\,($k$)
 & Orig. (\%) & CAD (\%) & OOD (\%) & Avg. & Orig. (\%) & CAD (\%) & OOD (\%) & Avg.   \\ 
\midrule
Original (3.4k)(0) & 90.7$_{\pm0.6}$ & 78.8$_{\pm1.7}$ & 80.6$_{\pm2.4}$ & 83.4 & 90.1$_{\pm0.5}$ & 76.8$_{\pm1.0}$ & 79.1$_{\pm2.1}$ & 82.0  \\ 
\midrule 
Weighted \ (1.7k)\,(16) & 89.2$_{\pm0.8}$ & 81.1$_{\pm1.3}$ & 82.9$_{\pm2.1}$ & 84.4 & 81.7$_{\pm3.0}$ & 72.0$_{\pm4.7}$ & 76.3$_{\pm6.6}$ & 76.7  \\ 
Paired \ (16)\,(16) & 86.9$_{\pm1.3}$ & 77.9$_{\pm2.2}$ & 83.9$_{\pm4.2}$ & 82.9 & 87.0$_{\pm1.4}$ & 77.4$_{\pm2.2}$ & 84.4$_{\pm3.5}$ & 82.9  \\ 
\midrule 
\multicolumn{9}{l}{\textbf{\citet{wang2021robustness}:} \ (1.7k)\,(0)} \\ 
- Pred.\ from top \ ($n''$=1,284) & 80.5 & 78.1 & 66.6 & 75.1 & 83.6 & 83.4 & 73.4 & 80.1 \\ 
- Ann.\ from top \ ($n''$=1,618) & 77.9 & 82.4 & 65.6 & 75.3 & 81.8 & 86.1 & 71.2 & 79.7 \\ 
- Ann.\ from all \ ($n''$=1,694) & 80.5 & 81.6 & 68.5 & 76.9 & 85.7 & 89.8 & 75.6 & 83.7 \\ 
\midrule 
\multicolumn{9}{l}{\textbf{Our models:} \ (1.7k)\,(16)} \\ 
- Mean Offset  & 87.7$_{\pm1.3}$ & 84.5$_{\pm2.0}$ & 80.5$_{\pm4.4}$ & 84.2 & 88.1$_{\pm1.2}$ & 85.6$_{\pm1.1}$ & 83.0$_{\pm3.3}$ & \textbf{85.6}  \\ 
- Mean Offset + Regression & 88.5$_{\pm1.0}$ & 84.7$_{\pm1.8}$ & 82.0$_{\pm4.0}$ & 85.1 & 88.3$_{\pm1.0}$ & 85.2$_{\pm1.5}$ & 83.4$_{\pm3.3}$ & \textbf{85.6}  \\ 
\bottomrule
\end{tabular}
\caption{SRoBERTa\textsubscript{\texttt{large}}}
\label{table:sroberta-large}
\end{table*}

\begin{table*}
\small
\centering
\begin{tabular}{lcccccccc}
\toprule
\multicolumn{9}{c}{SimCSE-RoBERTa\textsubscript{\texttt{large}}}\\ 
& \multicolumn{4}{c}{\textbf{Free Reg.:} $\lambda \in \lbrace 10^{-3},10^{-2},\ldots, 10^3\rbrace$} & \multicolumn{4}{c}{\textbf{Strong Reg.:} $\lambda \in \lbrace 1, 10,\ldots, 10^3\rbrace$} \\ 
\cmidrule(lr){2-5} \cmidrule(lr){6-9}
 \textit{Model} $\ $ ($n$)\,($k$)
 & Orig. (\%) & CAD (\%) & OOD (\%) & Avg. & Orig. (\%) & CAD (\%) & OOD (\%) & Avg.   \\ 
\midrule
Original (3.4k)(0) & 89.6$_{\pm0.7}$ & 75.7$_{\pm1.2}$ & 74.6$_{\pm2.6}$ & 80.0 & 88.2$_{\pm0.6}$ & 73.9$_{\pm0.9}$ & 73.9$_{\pm2.0}$ & 78.7  \\ 
\midrule 
Weighted \ (1.7k)\,(16) & 88.1$_{\pm0.8}$ & 78.5$_{\pm1.1}$ & 75.1$_{\pm2.3}$ & 80.6 & 77.5$_{\pm3.0}$ & 75.5$_{\pm3.5}$ & 72.8$_{\pm4.6}$ & 75.3  \\ 
Paired \ (16)\,(16) & 81.5$_{\pm2.2}$ & 80.9$_{\pm2.4}$ & 77.5$_{\pm4.3}$ & 80.0 & 81.2$_{\pm1.9}$ & 79.8$_{\pm3.0}$ & 78.0$_{\pm4.2}$ & 79.7  \\ 
\midrule 
\multicolumn{9}{l}{\textbf{\citet{wang2021robustness}:} \ (1.7k)\,(0)} \\ 
 - Pred.\ from top \ ($n''$=1,284) & 79.3 & 73.2 & 61.2 & 71.2 & 81.4 & 82.6 & 73.0 & 79.0 \\ 
- Ann.\ from top \ ($n''$=1,618) & 78.5 & 76.4 & 59.4 & 71.5 & 80.3 & 84.2 & 74.1 & 79.5 \\ 
- Ann.\ from all \ ($n''$=1,694) & 80.1 & 81.1 & 62.6 & 74.6 & 83.0 & 85.7 & 76.5 & 81.7 \\ 
\midrule 
\multicolumn{9}{l}{\textbf{Our models:} \ (1.7k)\,(16)} \\ 
 - Mean Offset & 87.0$_{\pm1.0}$ & 81.6$_{\pm1.4}$ & 74.1$_{\pm2.7}$ & 80.9 & 86.2$_{\pm1.2}$ & 84.6$_{\pm1.3}$ & 78.0$_{\pm3.2}$ & \textbf{83.0}  \\ 
- Mean Offset + Regression & 87.1$_{\pm1.4}$ & 82.5$_{\pm1.8}$ & 75.6$_{\pm3.1}$ & 81.7 & 86.1$_{\pm1.2}$ & 84.1$_{\pm1.3}$ & 78.2$_{\pm3.1}$ & 82.8  \\ 
\bottomrule
\end{tabular}
\caption{SimCSE-RoBERTa\textsubscript{\texttt{large}}}
\label{table:simcse-roberta-large}
\end{table*}

\begin{table*}
\small
\centering
\begin{tabular}{lcccccccc}
\toprule
\multicolumn{9}{c}{SDistilRoBERTa}\\ 
& \multicolumn{4}{c}{\textbf{Free Reg.:} $\lambda \in \lbrace 10^{-3},10^{-2},\ldots, 10^3\rbrace$} & \multicolumn{4}{c}{\textbf{Strong Reg.:} $\lambda \in \lbrace 1, 10,\ldots, 10^3\rbrace$} \\ 
\cmidrule(lr){2-5} \cmidrule(lr){6-9}
 \textit{Model} $\ $ ($n$)\,($k$)
 & Orig. (\%) & CAD (\%) & OOD (\%) & Avg. & Orig. (\%) & CAD (\%) & OOD (\%) & Avg.   \\ 
\midrule
Original (3.4k)(0) & 87.4$_{\pm0.8}$ & 78.2$_{\pm1.8}$ & 74.3$_{\pm3.7}$ & 80.0 & 87.2$_{\pm0.6}$ & 74.8$_{\pm0.8}$ & 72.7$_{\pm2.7}$ & 78.2  \\ 
\midrule 
Weighted \ (1.7k)\,(16) & 85.9$_{\pm0.8}$ & 80.3$_{\pm1.2}$ & 74.8$_{\pm3.2}$ & 80.3 & 77.1$_{\pm3.2}$ & 72.6$_{\pm3.2}$ & 72.6$_{\pm8.0}$ & 74.1  \\ 
Paired \ (16)\,(16) & 83.3$_{\pm2.0}$ & 78.0$_{\pm2.8}$ & 80.0$_{\pm5.6}$ & 80.4 & 83.7$_{\pm2.1}$ & 77.0$_{\pm2.9}$ & 80.7$_{\pm5.5}$ & 80.5  \\ 
\midrule 
\multicolumn{9}{l}{\textbf{\citet{wang2021robustness}:} \ (1.7k)\,(0)} \\ 
 - Pred.\ from top \ ($n''$=1,284) & 73.6 & 84.4 & 65.8 & 74.6 & 78.3 & 85.7 & 74.9 & 79.6 \\ 
- Ann.\ from top \ ($n''$=1,618) & 75.0 & 81.8 & 58.1 & 71.6 & 79.9 & 88.9 & 75.0 & 81.3 \\ 
- Ann.\ from all \ ($n''$=1,694) & 78.1 & 89.8 & 68.4 & 78.7 & 81.4 & 90.6 & 75.2 & 82.4 \\ 
\midrule 
\multicolumn{9}{l}{\textbf{Our models:} \ (1.7k)\,(16)} \\ 
 - Mean Offset & 83.4$_{\pm1.5}$ & 83.2$_{\pm2.4}$ & 74.2$_{\pm5.9}$ & 80.3 & 84.2$_{\pm1.2}$ & 85.8$_{\pm1.7}$ & 79.2$_{\pm4.8}$ & 83.0  \\ 
- Mean Offset + Regression & 84.1$_{\pm1.2}$ & 83.8$_{\pm2.1}$ & 75.7$_{\pm5.9}$ & 81.2 & 84.4$_{\pm1.2}$ & 85.1$_{\pm1.8}$ & 79.7$_{\pm4.7}$ & \textbf{83.1}  \\ 
\bottomrule
\end{tabular}
\caption{SDistilRoBERTa}
\label{table:SDistilRoBERTa}
\end{table*}

\begin{table*}
\small
\centering
\begin{tabular}{lcccccccc}
\toprule
\multicolumn{9}{c}{SMPNet}\\ 
& \multicolumn{4}{c}{\textbf{Free Reg.:} $\lambda \in \lbrace 10^{-3},10^{-2},\ldots, 10^3\rbrace$} & \multicolumn{4}{c}{\textbf{Strong Reg.:} $\lambda \in \lbrace 1, 10,\ldots, 10^3\rbrace$} \\ 
\cmidrule(lr){2-5} \cmidrule(lr){6-9}
 \textit{Model} $\ $ ($n$)\,($k$)
 & Orig. (\%) & CAD (\%) & OOD (\%) & Avg. & Orig. (\%) & CAD (\%) & OOD (\%) & Avg.   \\ \midrule
Original (3.4k)(0) & 90.3$_{\pm0.5}$ & 75.7$_{\pm1.4}$ & 78.9$_{\pm2.4}$ & 81.6 & 89.8$_{\pm0.5}$ & 72.3$_{\pm1.1}$ & 76.5$_{\pm1.9}$ & 79.5  \\ 
\midrule 
Weighted \ (1.7k)\,(16) & 88.9$_{\pm0.8}$ & 77.7$_{\pm1.5}$ & 79.1$_{\pm2.6}$ & 81.9 & 78.5$_{\pm3.0}$ & 67.9$_{\pm4.0}$ & 71.3$_{\pm7.4}$ & 72.6  \\ 
Paired \ (16)\,(16) & 84.4$_{\pm2.4}$ & 75.2$_{\pm3.0}$ & 76.7$_{\pm6.9}$ & 78.8 & 83.8$_{\pm3.0}$ & 73.9$_{\pm3.7}$ & 78.9$_{\pm6.3}$ & 78.9  \\ 
\midrule 
\multicolumn{9}{l}{\textbf{\citet{wang2021robustness}:} \ (1.7k)\,(0)} \\ 
 - Pred.\ from top \ ($n''$=1,284) & 78.7 & 73.6 & 65.6 & 72.6 & 83.0 & 83.6 & 76.4 & 81.0 \\ 
- Ann.\ from top \ ($n''$=1,618) & 80.5 & 82.6 & 70.0 & 77.7 & 81.6 & 86.7 & 75.5 & 81.2 \\ 
- Ann.\ from all \ ($n''$=1,694) & 83.2 & 88.7 & 76.5 & 82.8 & 83.4 & 88.7 & 79.2 & \textbf{83.8} \\ 
\midrule 
\multicolumn{9}{l}{\textbf{Our models:} \ (1.7k)\,(16)} \\ 
 - Mean Offset & 86.7$_{\pm1.7}$ & 82.2$_{\pm1.5}$ & 78.3$_{\pm3.5}$ & 82.4 & 86.8$_{\pm1.6}$ & 83.1$_{\pm1.5}$ & 79.8$_{\pm3.8}$ & 83.2  \\ 
- Mean Offset + Regression & 87.3$_{\pm1.5}$ & 82.5$_{\pm1.5}$ & 78.7$_{\pm3.4}$ & 82.8 & 87.3$_{\pm1.5}$ & 82.8$_{\pm1.5}$ & 79.9$_{\pm3.5}$ & 83.3  \\ 
\bottomrule
\end{tabular}
\caption{SMPNet}
\label{table:smpnet}
\end{table*}

\begin{table*}
\small
\centering
\begin{tabular}{lcccccccc}
\toprule
\multicolumn{9}{c}{SimCSE-BERT\textsubscript{\texttt{large}}}\\ 
& \multicolumn{4}{c}{\textbf{Free Reg.:} $\lambda \in \lbrace 10^{-3},10^{-2},\ldots, 10^3\rbrace$} & \multicolumn{4}{c}{\textbf{Strong Reg.:} $\lambda \in \lbrace 1, 10,\ldots, 10^3\rbrace$} \\ 
\cmidrule(lr){2-5} \cmidrule(lr){6-9}
 \textit{Model} $\ $ ($n$)\,($k$)
 & Orig. (\%) & CAD (\%) & OOD (\%) & Avg. & Orig. (\%) & CAD (\%) & OOD (\%) & Avg.   \\ 
\midrule
Original (3.4k)(0) & 88.6$_{\pm0.8}$ & 80.0$_{\pm1.1}$ & 80.1$_{\pm2.7}$ & 82.9 & 88.0$_{\pm0.4}$ & 80.3$_{\pm0.8}$ & 82.5$_{\pm1.4}$ & 83.6  \\ 
\midrule 
Weighted \ (1.7k)\,(16) & 87.2$_{\pm0.8}$ & 81.5$_{\pm1.2}$ & 81.9$_{\pm1.8}$ & 83.5 & 82.4$_{\pm1.4}$ & 84.2$_{\pm1.7}$ & 81.0$_{\pm3.3}$ & 82.5  \\ 
Paired \ (16)\,(16) & 83.1$_{\pm1.5}$ & 84.1$_{\pm3.3}$ & 83.8$_{\pm2.8}$ & 83.7 & 83.3$_{\pm1.5}$ & 83.8$_{\pm3.2}$ & 84.3$_{\pm2.3}$ & 83.8  \\ 
\midrule 
\multicolumn{9}{l}{\textbf{\citet{wang2021robustness}:} \ (1.7k)\,(0)} \\ 
- Pred.\ from top \ ($n''$=1,284) & 74.4 & 80.5 & 67.1 & 74.0 & 82.4 & 89.1 & 79.9 & 83.8 \\
- Ann.\ from top \ ($n''$=1,618) & 72.5 & 82.8 & 70.8 & 75.4 & 80.3 & 90.0 & 82.5 & 84.3 \\ 
- Ann.\ from all \ ($n''$=1,694) & 74.4 & 86.1 & 74.1 & 78.2 & 83.2 & 90.4 & 83.0 & 85.5 \\ 
\midrule 
\multicolumn{9}{l}{\textbf{Our models:} \ (1.7k)\,(16)} \\ 
 - Mean Offset & 85.7$_{\pm1.2}$ & 84.1$_{\pm1.9}$ & 80.6$_{\pm3.6}$ & 83.5 & 85.4$_{\pm0.8}$ & 87.1$_{\pm1.2}$ & 84.4$_{\pm1.7}$ & \textbf{85.6}  \\ 
- Mean Offset + Regression & 85.9$_{\pm1.0}$ & 85.1$_{\pm1.7}$ & 82.7$_{\pm2.5}$ & 84.6 & 85.5$_{\pm0.7}$ & 87.0$_{\pm1.1}$ & 84.4$_{\pm1.6}$ & \textbf{85.6}  \\ 
\bottomrule
\end{tabular}
\caption{SimCSE-BERT\textsubscript{\texttt{large}}}
\label{table:simcse-bert-large}
\end{table*}

\begin{table*}
\small
\centering
\begin{tabular}{lcccccccc}
\toprule
\multicolumn{9}{c}{SimCSE-BERT\textsubscript{\texttt{base}}}\\ 
& \multicolumn{4}{c}{\textbf{Free Reg.:} $\lambda \in \lbrace 10^{-3},10^{-2},\ldots, 10^3\rbrace$} & \multicolumn{4}{c}{\textbf{Strong Reg.:} $\lambda \in \lbrace 1, 10,\ldots, 10^3\rbrace$} \\ 
\cmidrule(lr){2-5} \cmidrule(lr){6-9}
 \textit{Model} $\ $ ($n$)\,($k$)
 & Orig. (\%) & CAD (\%) & OOD (\%) & Avg. & Orig. (\%) & CAD (\%) & OOD (\%) & Avg.   \\ 
\bottomrule
Original (3.4k)(0) & 88.8$_{\pm0.8}$ & 77.9$_{\pm1.1}$ & 78.9$_{\pm2.4}$ & 81.9 & 88.2$_{\pm0.5}$ & 74.9$_{\pm0.9}$ & 77.0$_{\pm1.7}$ & 80.0  \\ 
\midrule 
Weighted \ (1.7k)\,(16) & 87.4$_{\pm1.0}$ & 79.7$_{\pm1.4}$ & 77.9$_{\pm2.5}$ & 81.7 & 79.0$_{\pm2.6}$ & 77.7$_{\pm2.9}$ & 75.4$_{\pm5.9}$ & 77.3  \\ 
Paired \ (16)\,(16) & 80.8$_{\pm2.5}$ & 80.6$_{\pm3.3}$ & 79.4$_{\pm4.2}$ & 80.3 & 80.9$_{\pm2.2}$ & 79.8$_{\pm3.3}$ & 81.1$_{\pm3.2}$ & 80.6  \\ 
\midrule 
\multicolumn{9}{l}{\textbf{\citet{wang2021robustness}:} \ (1.7k)\,(0)} \\ 
 - Pred.\ from top \ ($n''$=1,284) & 80.3 & 77.5 & 70.9 & 76.2 & 83.6 & 79.9 & 74.7 & 79.4 \\
- Ann.\ from top \ ($n''$=1,618) & 80.5 & 82.0 & 72.2 & 78.2 & 82.6 & 84.8 & 77.4 & 81.6 \\ 
- Ann.\ from all \ ($n''$=1,694) & 82.0 & 82.8 & 75.1 & 79.9 & 82.0 & 86.1 & 80.9 & \textbf{83.0} \\ 
\midrule 
\multicolumn{9}{l}{\textbf{Our models:} \ (1.7k)\,(16)} \\ 
 - Mean Offset & 85.7$_{\pm1.5}$ & 83.1$_{\pm1.4}$ & 74.4$_{\pm3.5}$ & 81.1 & 84.9$_{\pm1.2}$ & 85.4$_{\pm1.1}$ & 78.8$_{\pm2.7}$ & \textbf{83.0}  \\ 
- Mean Offset + Regression & 86.1$_{\pm1.2}$ & 83.5$_{\pm1.4}$ & 76.6$_{\pm3.7}$ & 82.1 & 84.9$_{\pm1.2}$ & 84.8$_{\pm1.4}$ & 79.0$_{\pm2.9}$ & 82.9  \\ 
\bottomrule
\end{tabular}
\caption{SimCSE-BERT\textsubscript{\texttt{base}}}
\label{table:simcse-bert-base}
\end{table*}

\end{document}